\documentclass[11pt]{article}
\usepackage{amsmath}

\usepackage{longtable}
\usepackage{graphicx}
\usepackage{subcaption}
\usepackage[]{acl}
\usepackage{graphicx}
\usepackage{times}
\usepackage{booktabs}
\usepackage{multirow}
\usepackage{latexsym}
\usepackage{paralist}% For proper rendering and hyphenation of words containing Latin characters (including in bib files)
\usepackage[T1]{fontenc}
\usepackage{enumitem}
\usepackage{hyperref}
\usepackage{url}
\usepackage{amsmath}
\usepackage{amssymb}
\usepackage[utf8]{inputenc}

\usepackage{microtype}

\usepackage{inconsolata}

\usepackage{graphicx}

\title{A Comparative Analysis of LLM Memorization at Statistical and Internal Levels: Cross-Model Commonalities and Model-Specific Signatures}

\author{
Bowen Chen$^{1}$, Namgi Han$^{1}$, Yusuke Miyao$^{1,2}$ \\
$^1$Department of Computer Science, The University of Tokyo \\
$^2$Research and Development Center for Large Language Models, National Institute of Informatics\\
\texttt{\{bwchen, hng88, yusuke\}@is.s.u-tokyo.ac.jp}
}

\begin{document}
\maketitle
\begin{abstract}
Memorization is a fundamental component of intelligence for both humans and LLMs.
However, while LLM performance scales rapidly, our understanding of memorization lags.
Due to limited access to the pre-training data of LLMs, most previous studies focus on a single model series, leading to isolated observations among series, making it unclear which findings are general or specific.
In this study, we collect multiple model series (Pythia, OpenLLaMa, StarCoder, OLMo1/2/3) and analyze their shared or unique memorization behavior at both the statistical and internal levels, connecting individual observations while showing new findings.
At the statistical level, we reveal that the memorization rate scales log-linearly with model size, and memorized sequences can be further compressed.
Further analysis demonstrated a shared frequency and domain distribution pattern for memorized sequences. 
However, different models also show individual features under the above observations. 
At the internal level, we find that LLMs can remove certain injected perturbations, while memorized sequences are more sensitive. 
By decoding middle layers and attention head ablation, we revealed the general decoding process and shared important heads for memorization.
However, the distribution of those important heads differs between families, showing a unique family-level feature.
Through bridging various experiments and revealing new findings, this study paves the way for a universal and fundamental understanding of memorization in LLM.

\end{abstract}

\section{Introduction}
Large Language Models (LLMs) show unprecedented performance and have revolutionized the field of Natural Language Processing in recent years \cite{openai2025gpt5systemcard, anthropic2025claudesonnet45}. 
The progress in improving the model structure \cite{zhang2025mixtureexpertslargelanguage}, training algorithm \cite{guo2025deepseekr1}, and data \cite{zhou2025surveyllmtimesdata} is surprising, but the understanding of LLMs does not move equally.

While previous research analyzed LLM in many aspects \cite{yu-etal-2025-back,10.5555/3692070.3693122}, one of the mechanisms that is still not well-understood is \textit{Memorization} \cite{carlini2021extractingtrainingdatalarge,hartmann2023sokmemorizationgeneralpurposelarge,xiong2025landscapememorizationllmsmechanisms}, e.g., LLMs can regurgitate content in their pre-training corpora.
This special behavior makes LLMs a knowledge base \cite{Yang_2025}, or could lead to unfair usage in privacy or copyright issues \cite{chen2025surveyprivacyrisksprotection}.
While being a key element of intelligence in both humans and LLMs, the understanding of memorization is lacking since it requires access to training data, which largely restricts usable models.
Previous research usually focuses on a sandbox model or a single model series \cite{carlini2021extractingtrainingdatalarge,chen-etal-2024-multi-perspective, changalidis-harma-2025-capacity}, leading to isolated conclusions. 

This study aims to bridge this gap while revealing new findings by studying models with known pre-training data.
This includes 6 families (Pythia, StarCoder, OpenLLaMA, OLMo1, OLMo2, and OLMo3), for a total of 20 models, ranging from 1b to 32b, whose pre-training size spans from billion to trillion.
Though collecting sequences from various domains in each family, we compute the memorization score for each example in different model sizes.
Based on those examples, we comparatively analyze them from both the statistical and internal levels, aiming to connect the observations among individual models while bringing new findings. 
At the statistical level, we analyze the statistics of the collected scores.
We first study how the memorization rate couples with the model and data size, along with experiments showing the distribution of memorization scores of all examples and the least number of tokens to recall the same memorized sequence.
The following analysis discusses the distribution of memorized sequences over different domains and token frequency.
At the internal level, we inject noise into the model to analyze whether memorized sequences rely on a specific pattern or not.
To understand the emergence of memorized tokens, we also decode those memorized sequences in the middle layer.
Finally, to locate whether a memorization-specific pattern exists or not, we analyze the importance of each head by observing how generated sequences diverge from the memorized one without a specific head.
Our results show:
\begin{itemize}[leftmargin=*, noitemsep, topsep=0pt]

\item \textbf{Statistical Level:}
    \begin{itemize}[leftmargin=0.5em, noitemsep, topsep=0pt, label=\textendash]
        \item \textit{Cross-Model Commonalities:} Memorization rates universally scale log-linearly with parameter size. 
        Furthermore,  memorized sequences are intrinsically highly compressed (even using $\le 50\%$ of original tokens could obtain the same memorized output). 
        Additionally, most models shared a close memorization threshold for memorization, while larger models have a lower frequency threshold.
        At the domain level, most of the memorized sequences are structural texts (codes, etc), but scaling model size makes LLMs memorize more free texts.
        \item \textit{Model-Specific Signatures:} Stronger model capacity does not indicate a higher memorization rate. 
        Instead, we observe up to a 100$\times$ divergence in memorization rates across different families (e.g., StarCoder vs. OLMo1), heavily governed by their specific pre-training corpora.
    \end{itemize}
\item \textbf{Internal Level:}
    \begin{itemize}[leftmargin=0.5em, noitemsep, topsep=0pt, label=\textendash]
        \item \textit{Cross-Model Commonalities:} Under noise perturbation, LLMs reveal certain denoising capabilities that remove part of the injected noise, while the memorized sequences are harder to denoise. 
        Using attention ablation, we found that memorization-important heads highly overlap within domains, and a small subset of heads is important for most domains.
        \item \textit{Model-Specific Signatures:} While memorization important heads exist, their layer-wise distribution forms a distinct structural fingerprint.
        This distribution is significantly consistent within a model family, but diverges across different families, which is shaped by the individual training recipe of each family.
    \end{itemize}

\end{itemize}

\section{Related Work}
In this section, we review the literature on LLM memorization at both the statistical (analyzing through observing various LLM outputs) and internal levels (investigating internal components).

\subsection{Statistics Analysis of Memorization}
\citet{tirumala2022memorization}  analyzed memorization in LLM with a sandbox setting, showing that larger models memorize sequences more easily.
Following it, \citet{carlini2023quantifyingmemorizationneurallanguage} studied the pre-trained GPT-Neo \cite{gpt-neo} models, revealing that the number of memorized sequences is related to model capacity, prompt size, and the repetitions in the pre-training corpora.
\citet{chen-etal-2024-multi-perspective} showed that memorized sequences are shared across model sizes, and the token frequency gap between input and output is vital for generating memorized sequences.
To analyze memorization beyond fixed input, \citet{schwarzschild2024rethinkingllmmemorizationlens} developed the compression ratio, e.g, the maximum number of sequences that a certain length sequence can recall.
\citet{hayes-etal-2025-measuring} studied the memorized sequences under different sampling strategies and suggested that the actual memorization may be more than what we have observed.

\subsection{Internal Memorization Mechanism}
\citet{dai-etal-2022-knowledge} showed that the knowledge neuron exists in the later layers of BERT \cite{devlin-etal-2019-bert}.
However, the results from \citet{templeton2024scaling} suggested that the learned knowledge is represented as a vector rather than stored in individual parameters.
\citet{chen-etal-2024-multi-perspective} shows the varied decoding entropy feature for memorized and unmemorized sequences.
\citet{arnold-2025-memorization} analyzed the Intrinsic Dimensions (IDs) of memorized sequences in LLM and found that low ID complexity sequences are easy to memorize.
\citet{lasy-etal-2025-understanding} analyzed the internal mechanism of the GPT-Neo-125m model \cite{gpt-neo} by interrupting the input and observing whether the LLM can still output the memorized sequences, suggesting the existence of triggers and maintenance components for memorization.
\citet{changalidis-harma-2025-capacity} also analyzed the theoretical memorization capacity through sandbox experiments on real data.

This study stands on both statistical and internal perspectives, aiming to extend the analysis to a wider spectrum of model families.

\section{Experiments Setting}
We introduce the motivation for each experiment.
We first study the basic statistics of memorization, including its intercorrelation with multiple factors, its firmness measured by compression ratio, and the distribution of memorization scores (Section \ref{sec: statistics}). 
These statistics reveal general features alongside model-specific curves, serving for further discussion on domain and frequency distributions (Section \ref{sec: distribution over domain}, \ref{sec: frequency distribution}). 
The following studies examine noise robustness under internal perturbations to determine whether memorized sequences are sensitive (Section \ref{sec: robustness}).
We then dive into the layer-level decoding of memorized/unmemorized tokens to illustrate the general decoding process and specific features of memorized sequences (Section \ref{sec: logitlens}). 
Finally, the last sections discuss shared components based on attention ablation, analyzing their overlaps and structure (Sections \ref{sec: overlap of important heads}, \ref{sec: distribution of component},\ref{sec: distribution similarity of heads}).
\subsection{Model Setting}
 We use the base models of Pythia \cite{biderman2023pythiasuiteanalyzinglarge}, the OLMo1 \cite{groeneveld-etal-2024-olmo}, the OLMo2 \cite{olmo20252olmo2furious}, the OLMo3 \cite{olmo2025olmo3}, the OpenLLaMA \cite{openlm2023openllama}, and the StarCoder \cite{li2023starcoder} models, including the pre-trained models, general, and coding LLMs with details in the Table \ref{tab:model-family-size-summary}.
For each model, we sample 300,000 sequences for each domain, covering various text types.\footnote{We present domain details in the Appendix \ref{sec:exp detail}. The OLMo1 is originally called \href{https://huggingface.co/allenai/OLMo-7B}{OLMo}, we specifically use OLMo1 to represent OLMo to avoid confusion.}

\begin{table}[t]
\centering
\scriptsize
\setlength{\tabcolsep}{6pt}
\begin{tabular}{l l l}
\hline
\textbf{Model } & \textbf{Size} & \textbf{Pre-training Data} \\
\hline
\multirow{2}{*}{Pythia} & 160m, 410m, 1b & \multirow{2}{*}{Pile \cite{gao2020pile800gbdatasetdiverse}}\\
 & 2.8b, 6.9b, 12b & \\
 \hline

OLMo1 & 1b, 7b & Dolma \cite{dolma}\\
\hline
\multirow{2}{*}{OLMo2} & 1b, 7b, 13b & OLMo-Mix and Domino\\
 & 32b & \cite{olmo20252olmo2furious}\\
 \hline

OLMo3 & 7b, 32b & Dolma3 \cite{olmo2025olmo3} \\
\hline

OpenLLaMA & 3b, 7b, 13b & Redpajama \cite{weber2024redpajamaopendatasettraining}\\
\hline

StarCoder & 1b, 3b, 7b & The Stack \cite{Kocetkov2022TheStack}\\
\hline
\end{tabular}
\caption{Model families and their pre-training data.}
\label{tab:model-family-size-summary}
\end{table}

\subsection{Metrics and Methods}
\label{methods}

\paragraph{Memorization Score} We follow the memorization definition from \citet{274574}.
For an LLM, we prompt sampled context tokens $C_i= \{c_{(i,1)} \cdots c_{(i,n)} \}$ from its pre-training corpora $C$ and use greedy decoding to generate the predicted continuation tokens $X_i= \{ x_{(i,1)} \cdots x_{(i,n)} \}$.
We also collect the actual continuations $Y_i= \{ y_{(i,1)} \cdots y_{(i,n)} \}$ under this context.
The memorization score for sample $i$ is calculated as follows:
\begin{equation}
M_i(X, Y)=\dfrac{\sum_{k=1}^{n} \textbf{I}(x_{i,k} = y_{i,k})}{n}
\end{equation}
$n$ means the length of the continuation tokens.
$\textbf{I}$ is the Indicator Function.
If $M_i(X, Y) = 1$, the sequence is fully memorized under this context sequence. 
A sequence $Y$ is unmemorized if $M(X, Y) = 0$.  
We set both the number of input and output tokens as 32 \cite{chen-etal-2024-multi-perspective, biderman2023emergentpredictablememorizationlarge}, meaning a sequence is memorized if the following 32 continuation tokens can be fully recalled by the prompted 32 input tokens.

Additionally, we calculate \textbf{memorization rate} by dividing the number of memorized sequences by the number of whole sequences.
To further analyze those memorized sequences, we also implemented the \textbf{compression ratio} \cite{schwarzschild2024rethinkingllmmemorizationlens}, which is the least number of context tokens that also extracts the whole memorized sequences divided by the number of original context tokens.

\paragraph{Residual-Stream Relative Gaussian Noise Injection} is used to study the memorization robustness against noise perturbation in Section \ref{sec: robustness}.
At the first layer, let the residual-stream output tensor be $\mathbf H_\ell\in\mathbb R^{B\times T\times d}$. Noise injection is applied as
\begin{equation}
\tilde{\mathbf H}_\ell = \mathbf H_\ell + \boldsymbol\varepsilon,
\qquad
\boldsymbol\varepsilon\sim\mathcal N\!\left(\mathbf 0,\sigma_{\mathrm{eff}}^2\mathbf I\right).
\end{equation}
The standard deviation is
$
\sigma_{\mathrm{eff}}=
\alpha \cdot \operatorname{RMS}(\mathbf H_\ell)
$
where RMS is a scalar over all tensor elements:
\begin{equation}
\operatorname{RMS}(\mathbf H_\ell)
= \sqrt{\frac{1}{\lvert\mathbf H_\ell\rvert}\sum_u (\mathbf H_\ell)_u^2}
\end{equation}
The range of $\alpha$ is [0.0, 0.1, 0.2, 0.3, 0.4, 0.5].

\paragraph{Logit-Lens \cite{belrose2023eliciting}} is used to probe the internal decoding probability of LLMs in the Section \ref{sec: logitlens}. 
For a hidden state $\mathbf{h}_l$ at layer $l$, we apply the final layer normalization $\gamma$ and the unembedding projection $\mathbf{W}_U$ to compute the decoding probability over the vocabulary $V_l = \text{softmax}(\mathbf{W}_U \gamma(\mathbf{h}_l))$. 

\begin{figure*}[t]
    \centering
    % ---- top figure ----
    \begin{subfigure}{\textwidth}
\centering\includegraphics[width=\textwidth,height=0.36\textwidth]{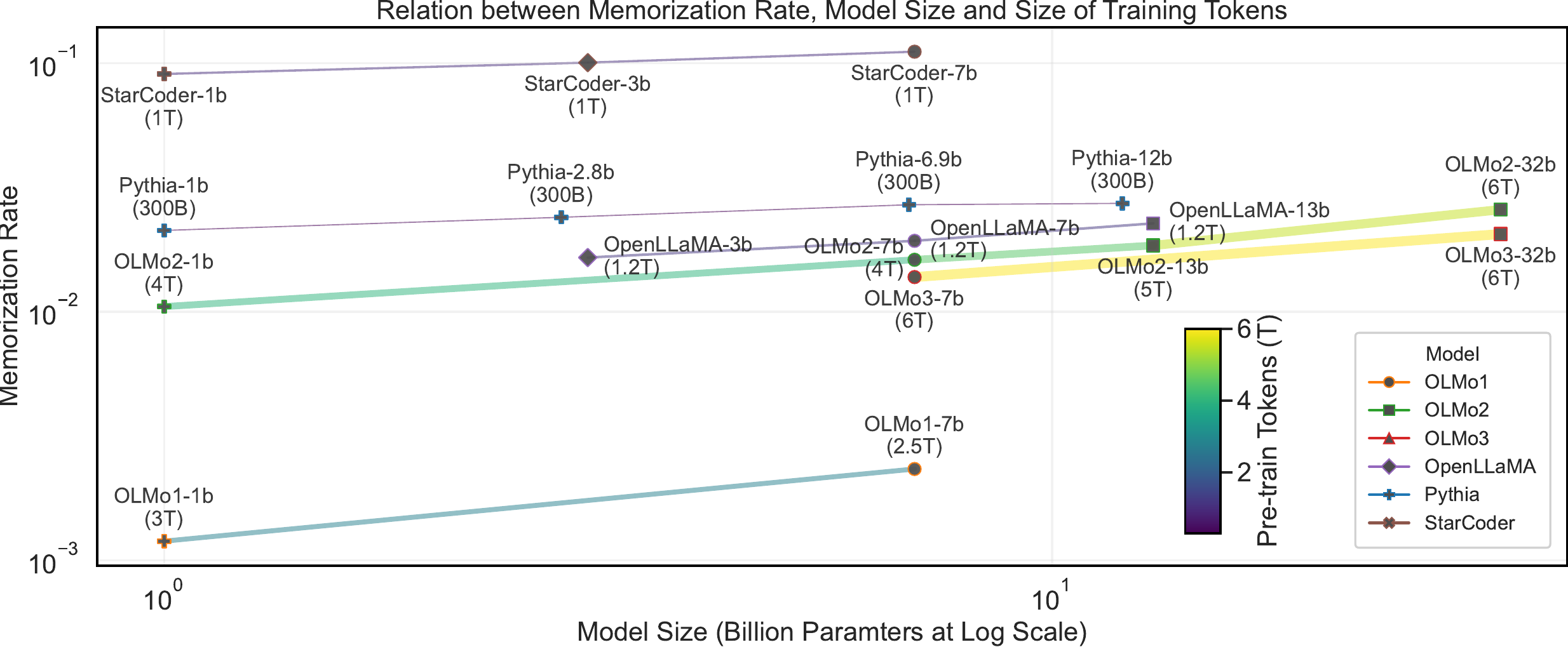}
        \caption{Memorization rate change with model size and training token size labeled for each model.}
        \label{fig: general_upper}
    \end{subfigure}
    \vspace{0em}
    \begin{subfigure}{0.38\textwidth}
            \centering
    \includegraphics[width=\textwidth]{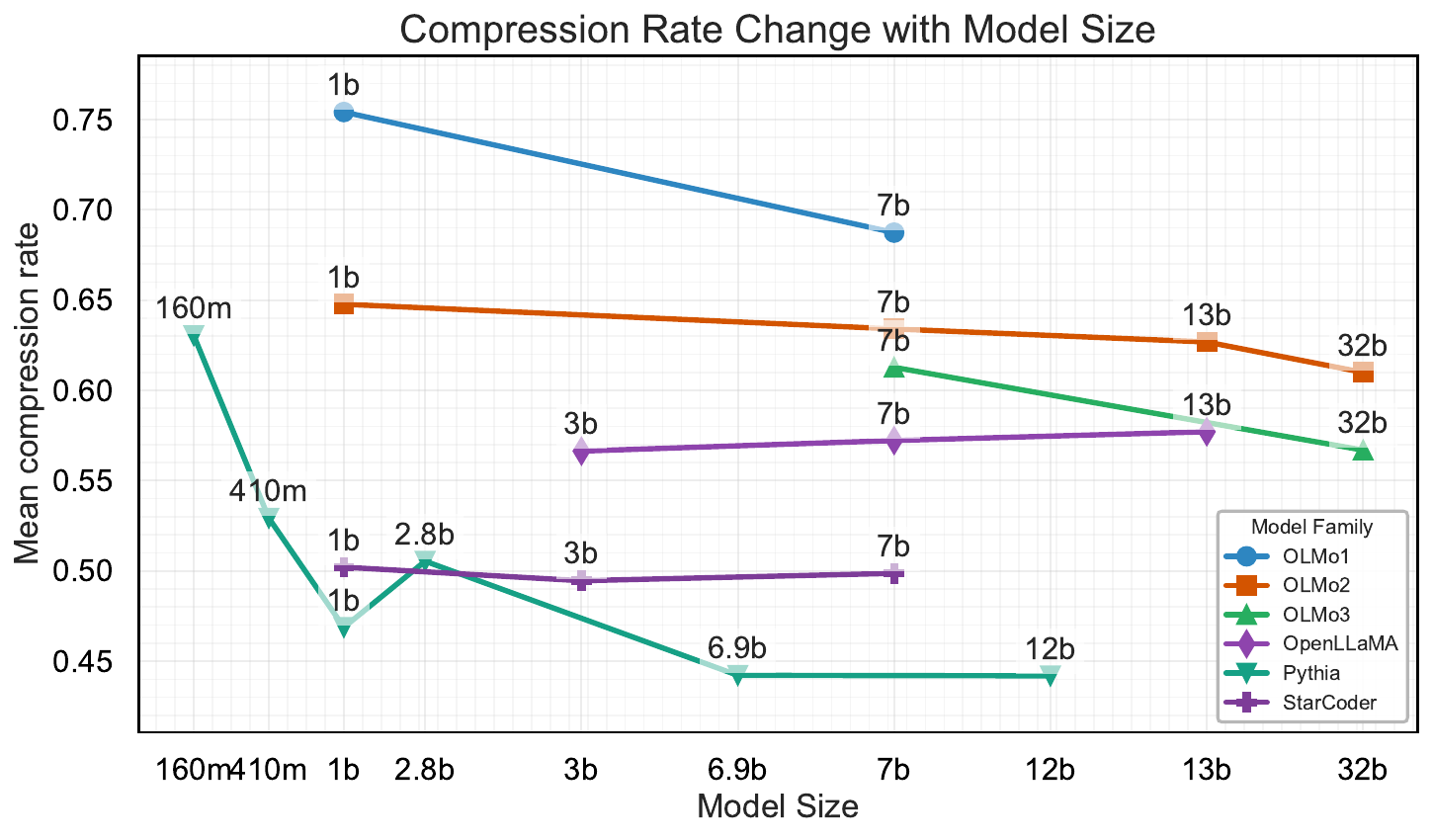}
            \caption{Compression rate across model sizes.}
            \label{fig: compression_rate}
        \end{subfigure}
    % ---- bottom left (60%) ----
    \begin{subfigure}{0.6\textwidth}
        \centering
        \includegraphics[width=\textwidth]{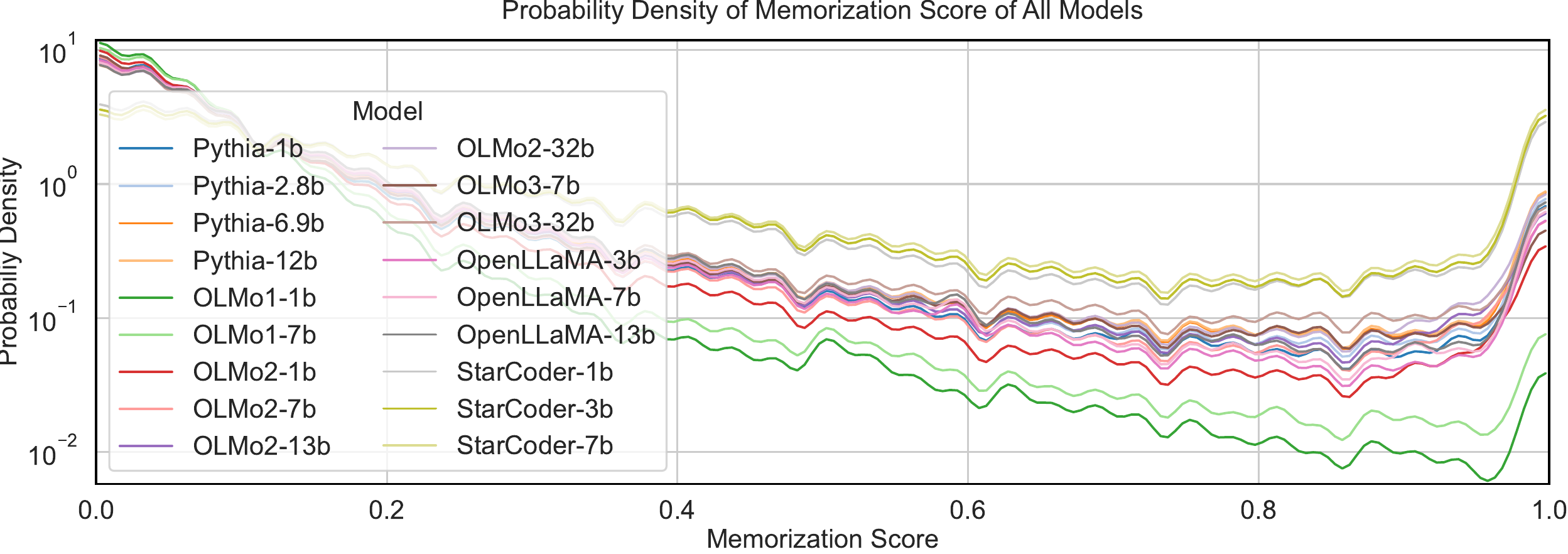}
        \caption{The distribution of memorization score of different models.}
        \label{fig:general_bottom}
    \end{subfigure}
    \hfill
    % ---- bottom right (40%) ----

    \caption{(Top Figure) The relationship between memorization rate, model size, and pre-training corpus size. The x/y-axis represents parameter counts and memorization rate on a log scale. Line width and color also encode the training token count (thicker lines = more tokens). 
    (Bottom left) The compression ratio across models. A smaller compression ratio requires fewer tokens to generate the same memorized sequences. (Bottom right) The probability density distribution of memorization scores for all models (log-scaled y-axis). }
    \label{fig: compression_rate_of_all}
\end{figure*}

\paragraph{Attention-Head Ablation and Importance} is used to study the head importance in Section \ref{sec: overlap of important heads}.
Let attention output at layer $\ell$ be $\mathbf A_\ell\in\mathbb R^{B\times T\times d}$, with $H$ heads.
For one head, we replace its value with the average of the other heads at this layer
$\frac{1}{H-1}\sum_{j\ne h}\mathbf A'_{\ell,:,:,j,:}$.
We can then get the generated sequence $P_i^{(\ell,h)}$ for this sample $i$ with head $h$ ablated from layer $l$.  
Let $T_i$ be the original generated memorized sequence without head ablation with $N$ tokens. 
The Head Importance Score $IM$ is
\begin{equation}
IM_{i,\ell,h}
= \frac{1}{N}\sum_{n=1}^{N}I(p_{i,k}==t_{i,k}^{(\ell,h)}\bigr),
\end{equation}
Larger $IM_{i,\ell,h}$ indicates head $(l,h)$ is important for generating sequence $i$.

% \paragraph{Distribution Shift Across Noise Levels (Wasserstein-1).}
% For each model $m$ and noise level $n$, scores are histogrammed on $[0,1]$ with $K$ bins (typically $K=L_r$), yielding PMF $p_{m,n}(k)$ and CDF
% \begin{equation}
% F_{m,n}(k)=\sum_{j=1}^{k}p_{m,n}(j),\qquad k=1,\dots,K.
% \end{equation}
% Within family $f$, average CDFs over models:
% \begin{equation}
% \bar F_{f,n}(k)=\frac{1}{\lvert M_f\rvert}\sum_{m\in M_f}F_{m,n}(k).
% \end{equation}
% Relative to baseline noise $n_0$, the implemented discrete 1D Wasserstein-1 distance is
% \begin{equation}
% W_1^{(f)}(n,n_0)
% = \Delta x\sum_{k=1}^{K}\left|\bar F_{f,n}(k)-\bar F_{f,n_0}(k)\right|,
% \qquad
% \Delta x=\frac{1}{K},
% \end{equation}
% which is the Riemann-sum form of $\int_0^1 \lvert F_n(x)-F_{n_0}(x)\rvert\,dx$.

\section{Results}
\subsection{Memorization Statistics}
\label{sec: statistics}
In this section, we begin by revealing the intercorrelation between memorization rate, model size, and data size.
For those memorized sequences, we also analyze their compression ratio \cite{schwarzschild2024rethinkingllmmemorizationlens} to study how firmly those memorized sequences are memorized.
Finally, we present the distribution of the collected memorization scores.
The results are presented in the Figure \ref{fig: compression_rate_of_all}.

We first observe that a larger model naturally enables more memorization, evidenced by a near-log-linear increase in the memorization rate as model size scales. 
However, this rate varies significantly across model families. 
StarCoder shows the highest rate (nearly 10\%), whereas OLMo1 shows the lowest since it is trained on a larger corpus than Pythia or StarCoder, while with a less advanced training algorithm compared to OLMo2/3.
Interestingly, a more capable model does not imply a higher memorization rate.
The OLMo3 has a lower rate than OLMo2 at the same sizes since it is trained on a larger corpus (6T tokens), but the \emph{absolute number} of memorized sequences still scales as memorization is close.\footnote{Detailed numbers are in the Table \ref{tab:memorized-domain-distribution}}
While pre-processing techniques like deduplication \cite{lee-etal-2022-deduplicating} help to reduce memorization, we see that advancements in architectural and training balances pre-processing, leading to a net increase in total memorized sequences.
\begin{figure*}[t] 
\centering % 让图像居中
\includegraphics[width=1\textwidth, height=0.43\textwidth]{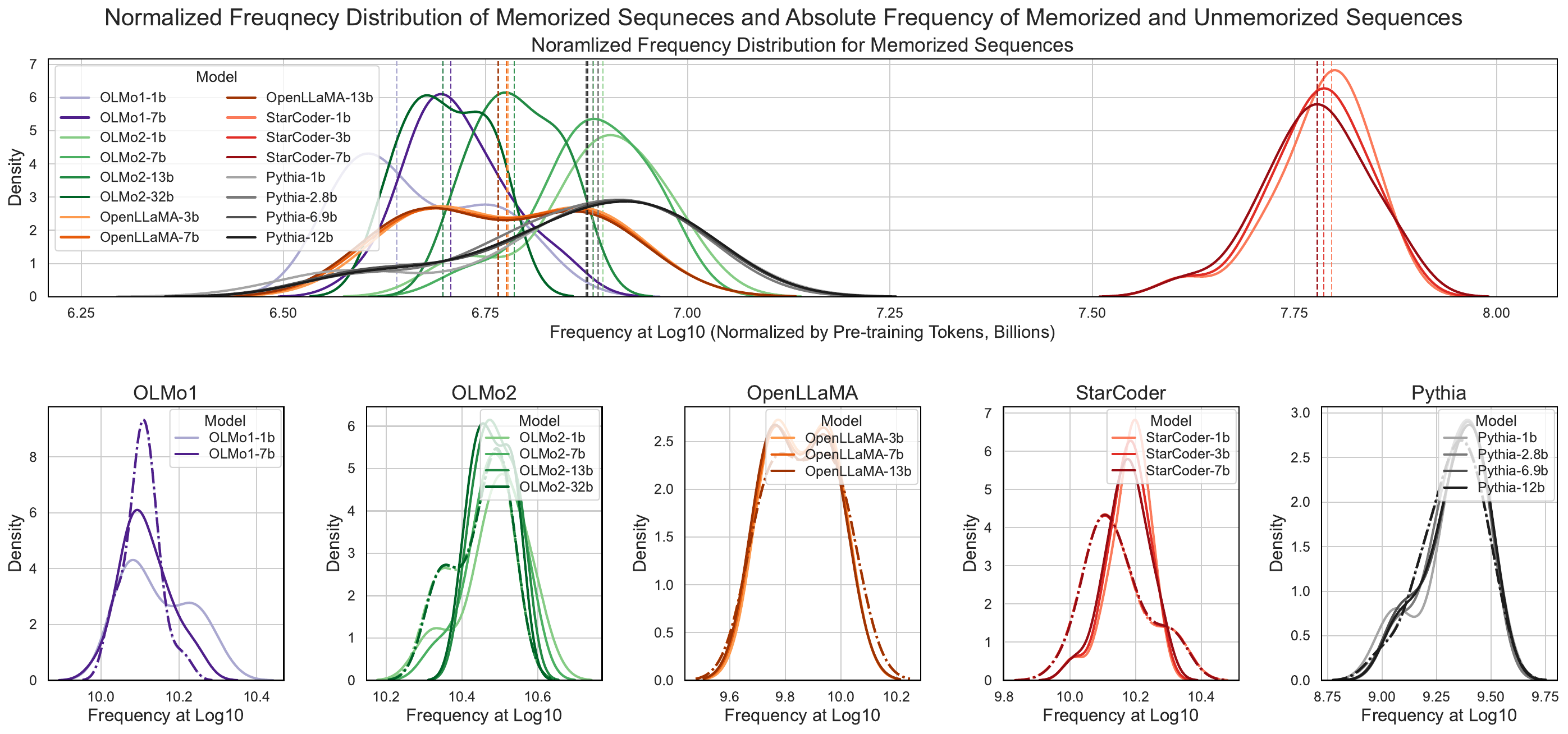}
\caption{The average token frequency distribution for different models. The upper figure shows the normalized frequency for memorized sequences. The dotted vertical lines indicate the average frequency for each model. The bottom figure shows the absolute frequency for memorized (solid line) and unmemorized (dotted line) sequences.}
\label{fig: memorization frequency}
\end{figure*}
\begin{table*}[t]
\centering
\resizebox{\textwidth}{!}{%
\begin{tabular}{lccccccccccccc}
\hline
Model & \multicolumn{3}{c}{Pythia} & \multicolumn{2}{c}{OLMo1} & \multicolumn{3}{c}{OLMo2} & \multicolumn{2}{c}{OLMo3} & \multicolumn{3}{c}{OpenLLaMA} \\
\cmidrule(lr){2-4} \cmidrule(lr){5-6} \cmidrule(lr){7-9} \cmidrule(lr){10-11} \cmidrule(lr){12-14}
Size & 1B & 2.8B & 12B & 1B & 7B & 7B & 13B & 32B & 7B & 32B & 3B & 7B & 13B \\
\hline
Structural & 75.1\% & 76.3\% & 74.3\% & 57.1\% & 49.7\% & 44.7\% & 43.9\% & 39.8\% & 49.0\% & 44.4\% & 65.9\% & 61.3\% & 58.7\% \\
Semi-Structural & 19.4\% & 17.2\% & 17.1\% & 32.3\% & 34.6\% & 9.7\% & 10.5\% & 12.1\% & 6.4\% & 7.9\% & 12.1\% & 13.5\% & 13.2\% \\
Free-Text & 5.5\% & 6.5\% & 8.6\% & 10.6\% & 15.7\% & 45.6\% & 45.6\% & 48.1\% & 44.7\% & 47.7\% & 21.9\% & 25.2\% & 28.1\% \\
\hline
\end{tabular}
}
\caption{Proportion of memorized texts across Structural, Semi-Structural, and Free-Text categories. StarCoder is not presented as it is only trained on code. }
\label{tab:memorized_category_proportions_noise0p0_layer0}
\end{table*}
For those memorized sequences, we also analyze their compression ratio in Figure \ref{fig: compression_rate}.
We see that the compression ratio generally decreases as model size scales. 
Pythia and StarCoder exhibit the lowest ratios (often $\le 0.5$), indicating highly compressed memory where less than half of the original context is sufficient to trigger a full memorized sequence. 
For most families (e.g., Pythia, OLMo1, OLMo2), the ratio drops consistently up to the 7B parameter. 
However, it often hits a plateau between 7B and 13B (notably in Pythia 6.9B–12B and OLMo2) before dropping again at the 32B scale. 
This plateau reveals a critical boundary in model scaling. While increasing parameters predictably expands the \emph{width} (total capacity) of memorization, the \emph{depth} (firmness) of these memories does not scale uniformly. 
Instead, even large models encounter bottlenecks in how firmly they can compress sequences, suggesting that memorization depth behaves more as an individual emergent property rather than a predictable byproduct of scaling.

Finally, the Figure \ref{fig:general_bottom} reveals a heavily bipolarized distribution of memorization scores across all models. 
The probability mass concentrates predominantly in the low-score region, paired with a sharp spike at the fully memorized area. 

\subsection{Distribution over Domains}
\label{sec: distribution over domain}
In this section, we show the distribution of memorized sequences over domains using three categories, Free/Semi-Structural/Structural texts, in the Table \ref{tab:memorized_category_proportions_noise0p0_layer0}. 
Free texts mean free-form natural language texts.
Semi-Structural texts mean language that has certain styles, like law documents.
Structural texts mean texts that strictly follow certain grammars, like codes or math expressions.
\footnote{For classification of each domain in different models, please refer to the Appendix \ref{sec: domain classification}. OLMo2 has a \emph{default} domain that contains mixed text types, and we treat it as free-text, which makes it have the highest proportion.}

The results show that for most models, structural texts occupy the largest proportion of memorized sequences, indicating they are easier to memorize.
Surprisingly, when we scale the model size, those LLMs do not further increase their memorization in the structural texts, but memorize more semi-structured and especially free texts.
This may be that the scaled model capability leads to an increase in language understanding, which eventually helps those LLMs to form semantic memorization that helps to memorize natural language.

\begin{figure*}[t]
\centering
\begin{subfigure}[t]{0.49\textwidth}
    \centering
    \includegraphics[width=\linewidth, height=0.5\linewidth]{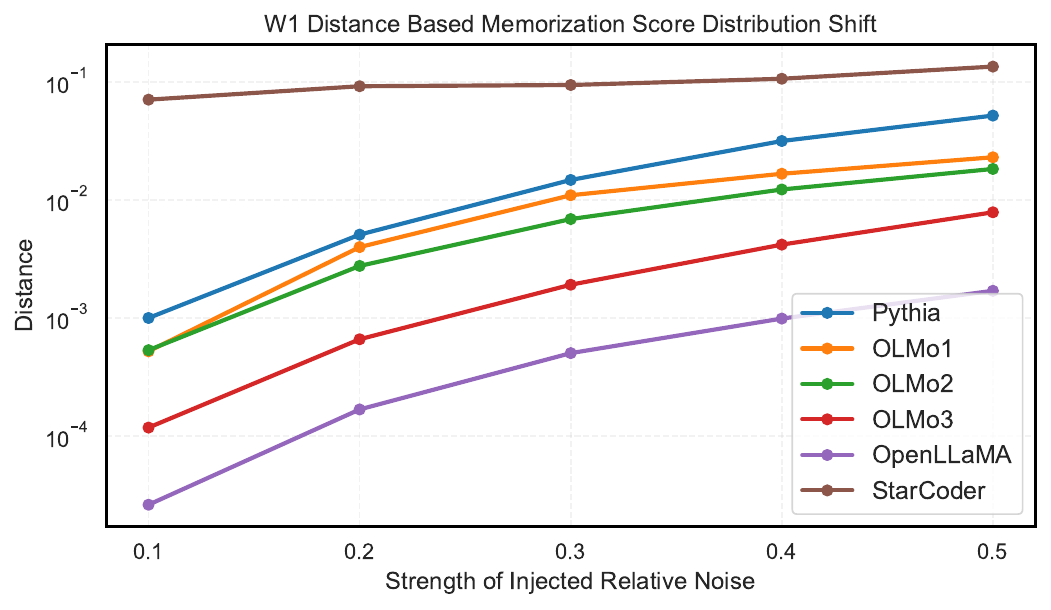}
    \caption{W1 distribution distance between noised and clean models.}
    \label{fig: general results}
\end{subfigure}
\begin{subfigure}[t]{0.49\textwidth}
    \centering
\includegraphics[width=\linewidth,height=0.5\linewidth]{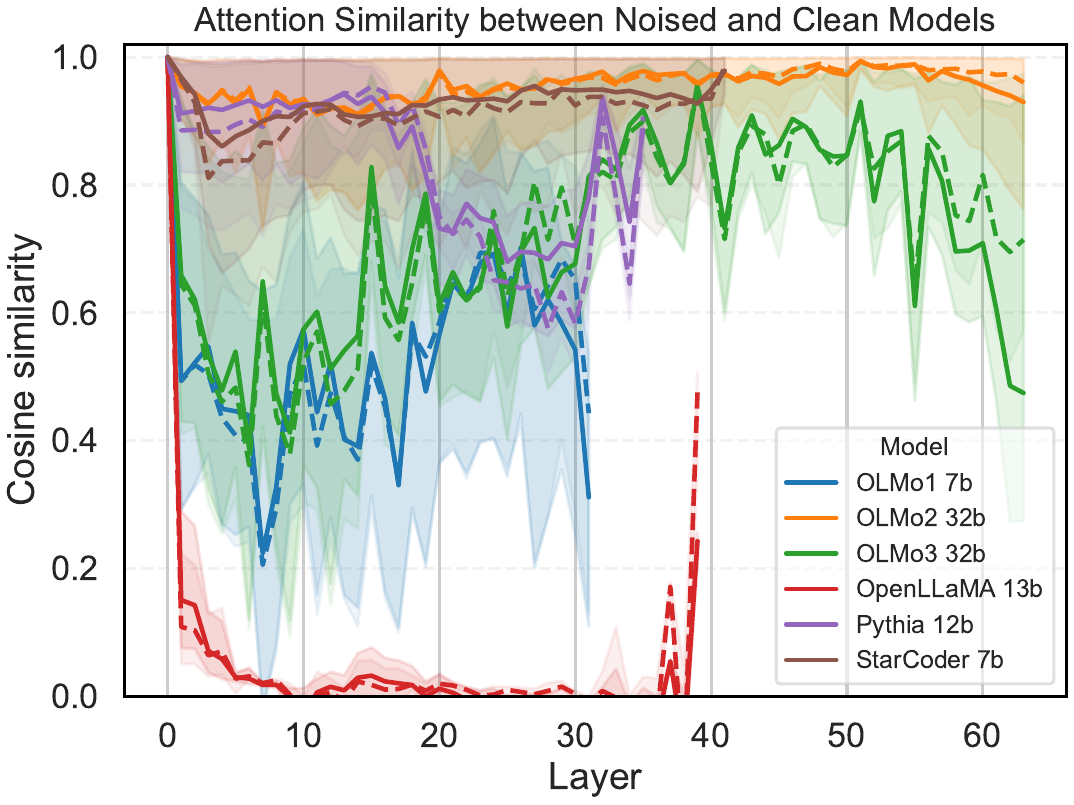}
    \caption{Attention similarity between noised and clean models. }
    \label{fig: component similarity}
\end{subfigure}
\caption{(Left figure) Memorization score distribution shift under noise. (Right figure) The Attention Head similarity between clean and noised models, with variance represented by the shadow. Solid/dotted lines represent results for memorized/unmemorized sequences. The noise range is [0.1, 0.2, 0.3, 0.4, 0.5]. }
\label{fig:w1_mlp_side_by_side}
\end{figure*}
\subsection{Frequency Distribution}
\label{sec: frequency distribution}

We show the average token frequency distribution for memorized and unmemorized sequences using Infini-gram \cite{Liu2024InfiniGram} in the Figure \ref{fig: memorization frequency}.\footnote{For detailed calculation method, please refer to Appendix Section \ref{sec: frequency}. The Infini-gram does not provide a query API for OLMo3 models.}

From the upper figure, we can first observe that, except for StarCoder, which is trained on a code-only domain, other models have an overlapped normalized frequency for memorized tokens, indicating a general frequency threshold for memorizing certain sequences.
Through observing their average frequency line, we see that they left-shift when scaling the model size in a model family when the training tokens are fixed, explaining the reason why large models memorize more sequences.
From the bottom figure, under the frequency overlap, we could see that unmemorized sequences have a higher distribution mass in the left area compared to memorized sequences, indicating that most of the unmemorized sequences tend to have a lower frequency.
However, the existence of memorized low-frequency sequences and high-frequency unmemorized sequences also indicates some extent of randomness in memorization, which is also found in previous research \cite{chen-etal-2024-multi-perspective}.

\subsection{Memorization Robustness Under Noise }
\label{sec: robustness}
Previous sections discussed the statistical level convergences and divergences.
From this section, we transition to the discussion of internal components, where the most intriguing question is \emph{do specific components exist for memorization}.
We study it by observing the memorization change against noise perturbation with the hypothesis that if memorization relies on certain specific components, they should be more sensitive to noise.
At the statistical level, the W1 distance is used \cite{10.1307/mmj/1029003026} to measure the memorization score distribution shift under noise.
The cosine similarity of attention heads between noise and clean models is used to measure the internal change under noise.

As shown in Figure \ref{fig: general results}, increasing noise strength shifts the memorization score distribution across all models.
The StarCoder is the weakest to noise perturbation as its learned representation is too domain specific to be noise-robust.
While other models share a similar curve, for OLMo families, we saw that more capable models are more robust to noise perturbation, meaning that while capable models may not have a lower compression ratio, they memorize sequences more firmly at the internal level.
To understand the internal mechanics behind this robustness, we analyze the layer-wise cosine embedding similarity between the noised and clean models (Figure \ref{fig: component similarity}). 
While noise predictably disrupts the early layers, similarity generally recovers in later layers across all models, revealing an internal error-correction mechanism akin to denoising. 
Crucially, however, memorized sequences exhibit lower similarity recovery compared to unmemorized sequences, showing that memorized sequences are more sensitive to the perturbation. 
This suggests that although LLMs possess general error-correction capabilities regardless of being memorized or not, memorized sequences rely on more sensitive computational pathways. 
Once those pathways are perturbed, the denoising mechanism fails to reconstruct the exact sequence, suggesting the existence of specific components for memorization.

\begin{figure*}[t] 
\centering % 让图像居中
\includegraphics[width=1\textwidth, height=0.3\textwidth]{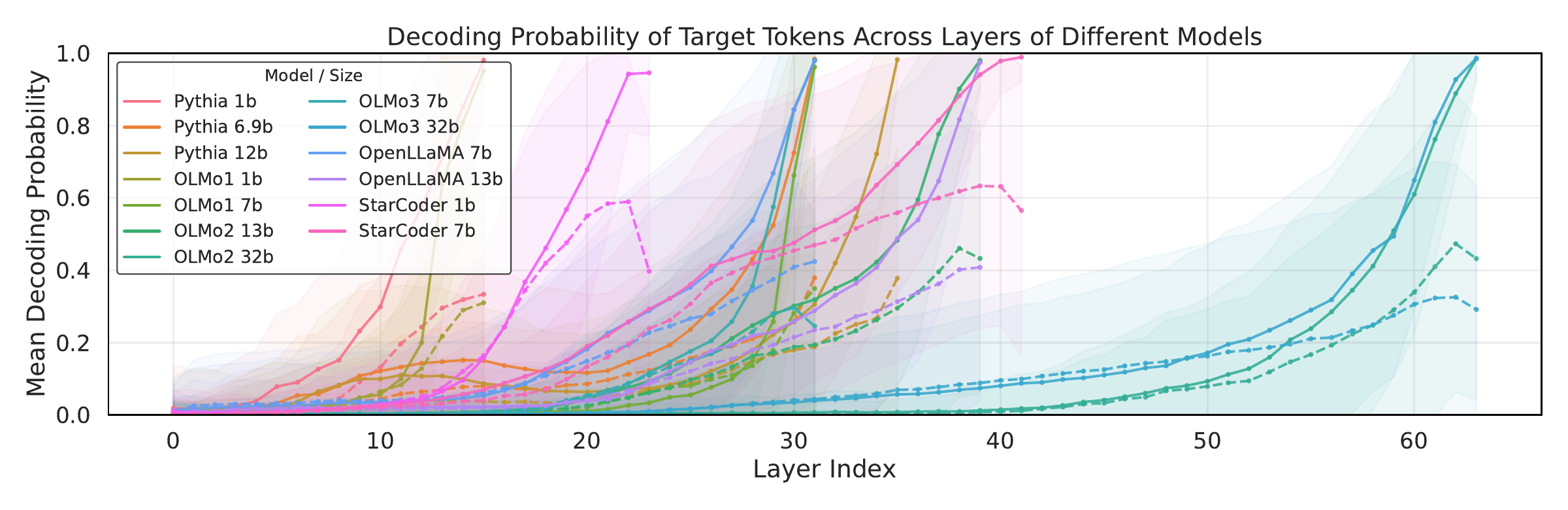}
\caption{Decoding probability (Y-axis) for memorized and unmemorized tokens across different layers (X-axis) for all models. The solid/dotted line represents the probability for memorized/unmemorized sequences.}
\label{fig: logit lens}
\end{figure*}
\subsection{Decoding the Middle Layers}
\label{sec: logitlens}
To further investigate the emergence of memorization, we apply the Logit-lens to probe token decoding probabilities across layers (Figure \ref{fig: logit lens}). 

\begin{table*}[ht]
\centering
\scriptsize
\setlength{\tabcolsep}{1.5pt}
\begin{tabular}{l l c c c c c c c c c}
\hline
\textbf{Model} & \textbf{Size} & \multicolumn{3}{c}{\textbf{Any Two Domain Pair}} & \multicolumn{1}{c}{\textbf{In-Domain Example Pair}} & \multicolumn{5}{c}{\textbf{Percentage of Shared Important Heads Among $>x\%$ Domains}} \\
\cmidrule(lr){3-5} \cmidrule(lr){6-6} \cmidrule(lr){7-11}
 & & \textbf{Mean Jaccard} & \textbf{Max Jaccard} & \textbf{Random} & \textbf{Mean Jaccard} & \textbf{$>30\%$ dom} & \textbf{$>50\%$ dom} & \textbf{$>70\%$ dom} & \textbf{$>90\%$ dom} & \textbf{all dom} \\
\hline
\multirow{2}{*}{OLMo1} & 1b & 28.58\% & 57.58\% & 11.39\% & 57.86\% & 28.12\% & 12.11\% & 5.08\% & 1.17\% & 0.78\% \\
 & 7b & 24.41\% & 48.01\% & 11.14\% & 80.03\% & 29.79\% & 10.84\% & 2.05\% & 0.98\% & 0.98\% \\
\hline
\multirow{4}{*}{OLMo2} & 1b & 52.08\% & 89.09\% & 11.39\% & 57.69\% & 23.05\% & 18.75\% & 14.06\% & 7.42\% & 4.69\% \\
 & 7b & 21.97\% & 70.12\% & 11.14\% & 82.26\% & 23.24\% & 9.96\% & 3.42\% & 0.78\% & 0.29\% \\
 & 13b & 18.93\% & 38.23\% & 11.12\% & 87.66\% & 20.62\% & 8.94\% & 1.75\% & 0.62\% & 0.38\% \\
 & 32b & 20.76\% & 25.49\% & 11.12\% & 91.18\% & 42.58\% & 14.30\% & 3.12\% & 3.12\% & 3.12\% \\
\hline
\multirow{2}{*}{OLMo3} & 7b & 22.51\% & 41.87\% & 11.14\% & 93.66\% & 20.31\% & 8.89\% & 2.64\% & 1.17\% & 0.29\% \\
 & 32b & 22.46\% & 27.05\% & 11.12\% & 96.57\% & 30.90\% & 16.17\% & 3.71\% & 1.37\% & 1.37\% \\
\hline
\multirow{3}{*}{OpenLLaMa} & 3b & 17.21\% & 30.47\% & 11.18\% & 74.97\% & 33.29\% & 15.62\% & 1.56\% & 0.84\% & 0.84\% \\
 & 7b & 18.04\% & 33.99\% & 11.14\% & 83.91\% & 32.81\% & 15.23\% & 1.86\% & 0.49\% & 0.49\% \\
 & 13b & 17.69\% & 29.03\% & 11.12\% & 83.87\% & 31.44\% & 15.50\% & 1.44\% & 0.44\% & 0.44\% \\
\hline
\multirow{4}{*}{Pythia} & 1b & 46.17\% & 92.59\% & 11.48\% & 58.03\% & 19.53\% & 18.75\% & 17.19\% & 1.56\% & 1.56\% \\
 & 2.8b & 17.03\% & 37.58\% & 11.14\% & 69.84\% & 16.89\% & 6.74\% & 1.76\% & 0.20\% & 0.10\% \\
 & 6.9b & 19.72\% & 39.46\% & 11.14\% & 65.98\% & 27.34\% & 11.43\% & 1.66\% & 0.39\% & 0.10\% \\
 & 12b & 14.97\% & 29.15\% & 11.13\% & 67.55\% & 16.25\% & 3.75\% & 0.76\% & 0.14\% & 0.00\% \\
\hline
\multirow{3}{*}{StarCoder} & 1b & 18.38\% & 67.39\% & 11.20\% & 94.79\% & 19.01\% & 6.51\% & 2.08\% & 0.52\% & 0.26\% \\
 & 3b & 15.69\% & 24.71\% & 11.19\% & 95.29\% & 19.44\% & 4.17\% & 1.01\% & 0.00\% & 0.00\% \\
 & 7b & 13.57\% & 19.03\% & 11.14\% & 96.84\% & 19.35\% & 1.04\% & 0.00\% & 0.00\% & 0.00\% \\
\hline
\end{tabular}
\caption{Cross-domain top-20\% head overlap against a random-selection, together with mean within-domain example-level overlap and the ratio of heads that remain top-20\% important in at least X\%, or all domains.}
\label{tab:head-overlap-random-significance}
\end{table*}

For both memorized and unmemorized sequences, decoding probabilities increase gradually in the early layers before exhibiting a sharp burst in the later layers. 
This shared trend suggests that the late-layer probability spike is primarily a general mechanism for output generation, rather than a phenomenon exclusive to memorization. 
Despite this shared pattern, memorized tokens maintain a distinct probability advantage (often $>0.5$) over unmemorized tokens in the later layers, becoming highly predictable well before the final layer. 
Linking this with our previous observation—where noise perturbation severely disrupts the internal similarity in early layers while late layers attempt to recover it—we hypothesize a two-stage mechanism for memorization: the core features and routing for exact memorization are likely established in the early layers, whereas the later layers primarily serve to activate and translate this prepared information into final output probabilities.

\subsection{Overlap of Important Heads}
\label{sec: overlap of important heads}
The results from noise robustness and layer decoding suggested that a shared common pattern exists in generating memorized sequences. 
To further locate them, we utilize Attention Head Ablation to trace the importance of each head for memorized examples in each domain.
Then, the importance score list of all heads at each example in different domains can be sorted and compared.
For each model, we analyze within and cross-domain overlap for the top 20\% important heads in Table \ref{tab:head-overlap-random-significance}.

First, in-domain example overlap is notably high, meaning most of the important heads are shared within examples of each domain, and this share increases with model size, suggesting that larger models compress domain-specific memorization into dedicated heads.
Conversely, cross-domain overlap is generally lower, meaning important heads in each domain do not overlap very much, and this overlap decreases as models scale, indicating that heads for different domains decouple in larger parameter spaces.
Despite this decoupling, for almost all models, a small but crucial subset of heads remains important for over 90\% (or even all) of the domains. 
For example, in the OLMo3 32b models, around 1.4\% of heads are important for all domains despite a large parameter space.
This indicates the existence of some heads that are crucial for the memorization behavior itself.

\subsection{Example Distribution of Shared Important Heads}
\label{sec: distribution of component}
Based on our discovery of cross-domain, universally important attention heads, we investigate their layer-wise distribution in the OLMo2-13B and Pythia-12B models because of the close layer numbers.
\begin{figure}[t] 
\centering % 让图像居中
\includegraphics[width=1\columnwidth,height=0.42\columnwidth]{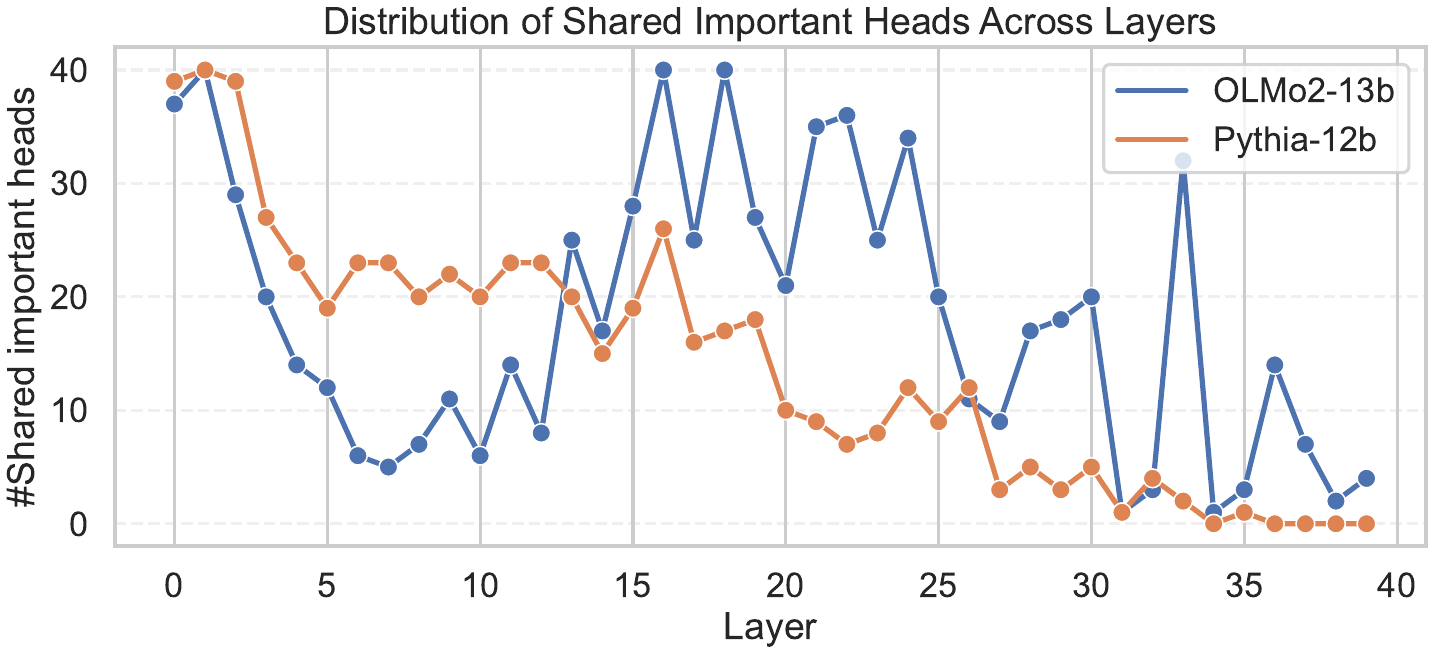}
\caption{Distribution of shared important heads across layers for Pythia-12b and OLMo-13b}
\label{fig: distribution_over_layers}
\end{figure}
As shown in Figure \ref{fig: distribution_over_layers}, these shared memorization heads do not distribute each layer uniformly but show distinct patterns based on models. 
While the general trend is that the most important heads for both models are in the early-middle layers, the distribution after the early-middle layers differs.
The OLMo2 models have a certain number of important heads also in the middle-later layers, while the numbers in Pythia stably decrease.
This fluctuating distribution further explains the results in Figure \ref{fig: logit lens}, where we hypothesize that the information is prepared at early layers but activated in the later layers.
In the early-middle layers, the general retrieval for memorization relies on shared important heads. 
As information progresses to deeper layers, shared heads activate domain-specific memorization coming from specialized heads, leading to a decrease in shared important heads. 
Finally, the exact memorized tokens are activated and reinforced by gathered information in the latest layers, leading to near-deterministic decoding.

\subsection{Distribution Similarity of Layer Importance Across Models}
\label{sec: distribution similarity of heads}
\begin{figure}[t] 
\centering % 让图像居中
\includegraphics[width=0.8\columnwidth]{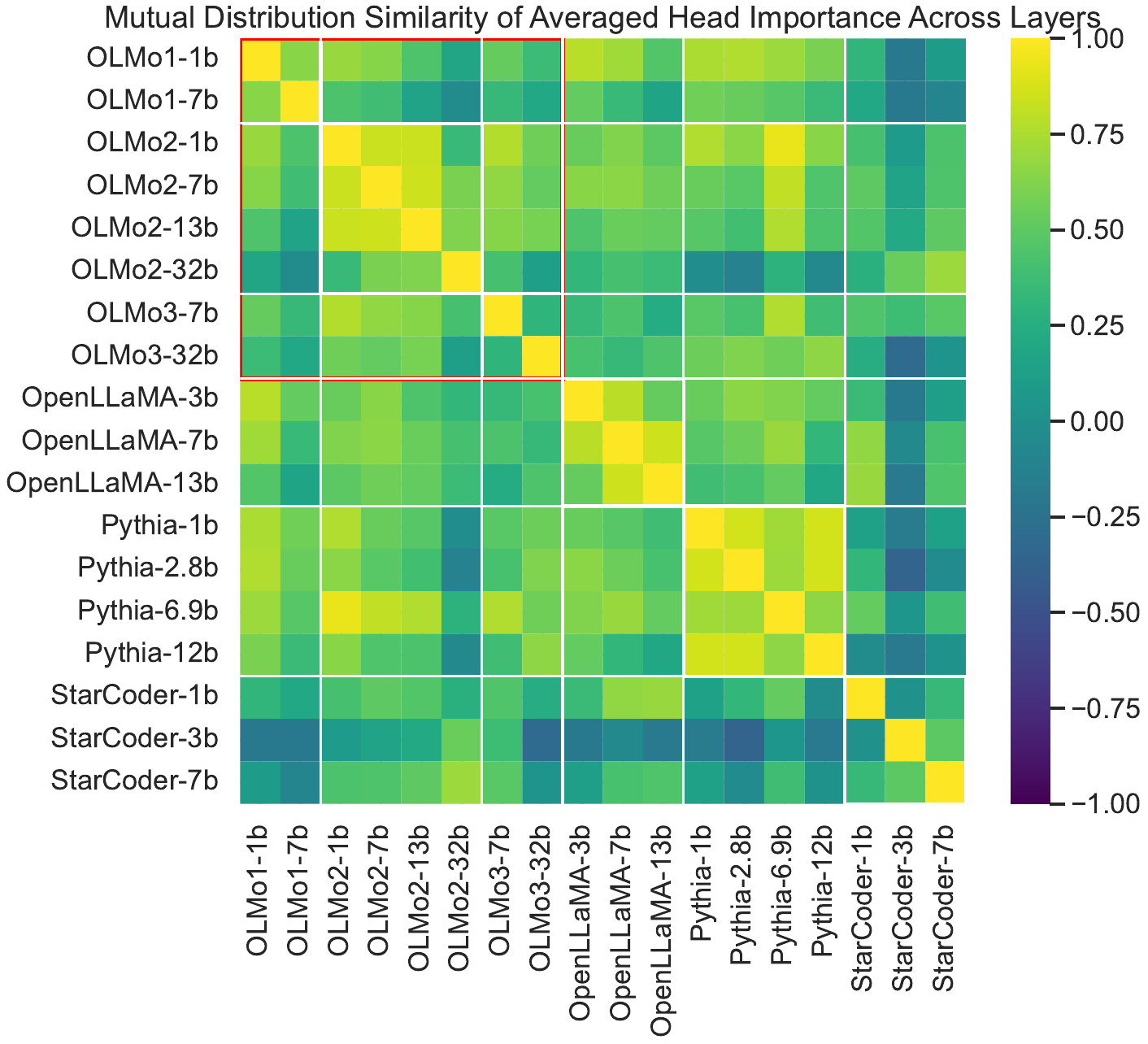}
\caption{Distribution Similarity Heatmap of Layer-wise Average Head Importance for all models.}
\label{fig: component heatmap}
\end{figure}

To understand whether different models share the same memorization structure or not, we compute a layer-wise importance score by taking the average of the importance scores of attention heads in each layer.
This layer-level importance is normalized based on the number of layers, with distribution comparison between each model in Figure \ref{fig: component heatmap}.

From this figure, we see a high similarity within the same model family. 
The Pythia has an average similarity close to 0.9 for each model size.
Noticeably, for OLMo1/2/3, not only at each model family, we even observed a cross-family similarity, indicating inherited training "DNA" within the temporally updated model series.
Additionally, different model families have a lower cross-similarity, especially the StarCoder, which has the farthest distance from other models.
This suggests that the memorization structure of a model family is shared among its models, regardless of size. 
However, there does not exist a universal memorization structure that exists for all LLMs, and the memorization structure is decided by the training recipe (model, data, training algorithm) of each model family.

\section{Conclusion}
In this study, we conducted a comparative analysis of LLM memorization at both the statistical and internal levels to identify cross-model commonalities and model-specific signatures. 
At the \textbf{statistical level}, despite individual model differences, we uncover a general log-linear scaling of memorization rate with model size, and show that memorized sequences are highly compressible. 
Further, the domain and frequency distribution show that large LLMs have a lower frequency threshold for memorization and tend to memorize more free texts instead of structural texts that are easy to memorize.
At the \textbf{internal level}, noise perturbation reveals that while LLMs show general denoising capability, memorized sequences are more sensitive, indicating a reliance on specific components. 
While the memorized sequence shows a stronger signal, decoding probability reveals a shared generation process. 
Finally, the attention ablation suggests that the memorization-important heads are shared within domains, and a small set of them is domain-irrelevant, indicating they may relate to the memorization itself. 
Crucially, the distribution of these heads is consistent within families but diverges across them, showing the memorization structure is shaped by each model's training recipe.
\section*{Limitations}
Though we have studied most of the commonly used fully open LLMs with data access, it is not practically feasible to use all existing models, since some models (Amber 7b, Apertus 70b, etc)\footnote{\url{https://github.com/LLM360/amber-data-prep}\\ \url{https://github.com/swiss-ai/pretrain-data}} release their pre-training data collection script rather than the data itself, meaning that the whole pre-training data collection and classification should be done by the user themselves.
This process is both heavy for the server and also does not guarantee the same collected data (invalid link, updated data, area access limitation, etc) while studying memorization requires the original training corpora. 
We selected models with clear data documentation or directly downloadable data through the Huggingface dataset or their individual links.
Additionally, most of the famous open models (Qwen, Deepseek, Llama, etc) do not release their pre-training data, so our experiments are not feasible for those LLMs.

Additionally, this research is conducted on the samples of the pre-training corpora, as the pre-training corpora are too huge to fully analyze in an economically friendly way, since even a storage device that could save their pre-training data would cost thousands of dollars per month, not even counting the computation cost.
However, we have made our best effort to validate our findings and results to be as general as possible, and the results also show common trends across model families and sizes within and across domains, indicating the current setting is trustworthy.

\section*{Ethical Considerations}
The usage of AI tools for this paper is limited to proofreading, and the usage of AI for the whole research includes coding assistance, and the results are validated by the authors.

While we have analyzed the sample pre-training data and sampled some examples for observation, we did not observe any offensive language or information related to personal privacy.
Additionally, for the sampling for the Pile pre-training data, we sample from its version where the copyright issues contents are removed\footnote{\url{https://huggingface.co/datasets/monology/pile-uncopyrighted}}.
Those domains include Books3, BookCorpus2, OpenSubtitles, YTSubtitles, and OWT2, and we did not sample data from those domains.

For all models and data used in this study, we align our usage with their license and intended usage.

\bibliography{custom}

\appendix

\section{Appendix}
\label{sec:appendix}
\subsection{Experiment Details}
\label{sec:exp detail}
\subsubsection{Device and Running Settings}
The running device of this research is mixed.
We have used H100, H200, A100 and A6000 servers.
In this study, we did not train any model, and basically, all CUDA devices are only used for generation and internal probing.
The running time depends on the model and its model size. 
To generate all memorization scores across 6 noise strengths for all models used in this study, it takes around 2 months for 8 A100 servers.

For the compression ratio analysis, it took around 2 weeks to run the experiments with 3 A6000 devices.

For the middle layer decoding analysis, it took around 2 weeks to run the experiments with 8 A100 for two weeks.

For the head importance results, it took around 1 month to run all model sizes in their domains on an H100 device.
The total number of attention heads grows very quickly with the increase of model size, since usually model size scaling is accompanied by layer number increase and expansion of head size, meaning that increasing the model size would lead to experimentation on additional hundreds of level attention heads.
Especially, we need to evaluate each head and regenerate the sequence without this head, which means one sequence requires an additional hundred times of generation when we use a larger model in a model family.
Therefore, we did not conduct experiments on all memorized examples in the head ablation study, as the computation cost is not practically affordable.
For small models (1b to 7b), we sample 10,000 memorized examples in each domain.
For large models (above 7b), we sample 2,500 memorized examples in each domain.

\subsubsection{Data Details}
\label{sec: domain classification}
We present the specific domains for each model and their classifications to the Free text, Semi-Structural, and Structural text categories in the Table \ref{tab:model-domain-classification-complete}.

For the collection of the data, we referred to either their released self-collected Huggingface dataset\footnote{\href{https://huggingface.co/datasets/allenai/dolma}{dolma data},
,\href{https://huggingface.co/datasets/allenai/olmo-mix-1124}{olmo-mix-1124 data},\href{https://huggingface.co/datasets/allenai/dolmino-mix-1124}{dolmino-mix-1124 data}
,\href{https://huggingface.co/datasets/allenai/dolma3_mix-6T}{dolma3 data}} or the dataset that was clearly mentioned in their posts or repository\footnote{\href{https://huggingface.co/datasets/bigcode/the-stack}{the stack data}

\href{https://huggingface.co/datasets/monology/pile-uncopyrighted}{pile data}

\href{https://huggingface.co/datasets/togethercomputer/RedPajama-Data-1T}{redpajama data}}.
We use a verified token to access that data and download it to the server, and separate it to the domain level.
After collecting the data, we pre-process them by passing those documents to their corresponding tokenizer of the target model and filtering the documents that have tokens less than 128, which we notice that only a very small amount of data is filtered by this process.
\begin{table*}[t]
\centering
\small
\setlength{\tabcolsep}{3pt}
\begin{tabular}{llll}
\hline
\textbf{Model Family} & \textbf{Domain Type} & \textbf{Domains} & \textbf{\#Dom} \\
\hline
\multirow{4}{*}{Pythia} & Structural & USPTO\_Backgrounds, EuroParl, Enron\_Emails, Github, StackExchange & 5 \\
& Semi-Structural & PubMed\_Abstracts, PubMed\_Central, FreeLaw, NIH\_ExPorter & 4 \\
& \multirow{2}{*}{Free-Text} & ArXiv, DM\_Mathematics, Gutenberg (PG-19), HackerNews, Pile-CC, & \multirow{2}{*}{8} \\
&  & PhilPapers, Ubuntu\_IRC, Wikipedia (en) &  \\
\hline
\multirow{5}{*}{OLMo1} & Structural & starcoder, proof\_pile\_2\_algebraic\_stack, proof\_pile\_2\_open\_web\_math & 3 \\
& \multirow{2}{*}{Semi-Structural} & redpajama\_stackexchange, cc\_news\_head, cc\_news\_middle, cc\_news\_tail, & \multirow{2}{*}{8} \\
&  & c4\_filtered, cc\_en\_head, cc\_en\_middle, cc\_en\_tail &  \\
& \multirow{2}{*}{Free-Text} & books, falcon\_refinedweb\_filtered, reddit, redpajama\_arxiv, & \multirow{2}{*}{8} \\
&  & wiki, pes2o, tulu\_flan, wikiref\_megawika &  \\
\hline
\multirow{3}{*}{OLMo2} & Structural & stackexchange, open-web-math, starcoder, math & 4 \\
& Semi-Structural & arxiv, pre\_train\_wiki, pre\_train\_dclm, pre\_train\_pes2o, pes2o & 5 \\
& Free-Text & default, flan, dclm & 3 \\
\hline
\multirow{19}{*}{OLMo3} & \multirow{4}{*}{Structural} & stack\_edu\_C, stack\_edu\_CSharp, stack\_edu\_Cpp, stack\_edu\_Go, stack\_edu\_Java, & \multirow{4}{*}{15} \\
&  & stack\_edu\_JavaScript, stack\_edu\_Markdown, stack\_edu\_PHP, stack\_edu\_Python, & \\
&  & stack\_edu\_Ruby, stack\_edu\_Rust, stack\_edu\_SQL, stack\_edu\_Shell, & \\
&  & stack\_edu\_Swift, stack\_edu\_TypeScript & \\
& \multirow{3}{*}{Semi-Structural} & dolma1\_7\_wiki\_en, finemath\_3plus, rpj\_proofpile\_arxiv, & \multirow{3}{*}{9} \\
&  & olmocr\_science\_pdfs\_adult\_content, olmocr\_science\_pdfs\_art\_and\_design, & \\
&  & olmocr\_science\_pdfs\_education\_and\_jobs, olmocr\_science\_pdfs\_entertainment
\\
&  & 
olmocr\_science\_pdfs\_sports\_and\_fitness & \\
& \multirow{12}{*}{Free-Text} & common\_crawl\_adult\_content, common\_crawl\_art\_and\_design, & \multirow{12}{*}{24} \\
&  & common\_crawl\_crime\_and\_law, common\_crawl\_education\_and\_jobs, & \\
&  & common\_crawl\_electronics\_and\_hardware, common\_crawl\_entertainment, & \\
&  & common\_crawl\_fashion\_and\_beauty, common\_crawl\_finance\_and\_business, & \\
&  & common\_crawl\_food\_and\_dining, common\_crawl\_games, & \\
&  & common\_crawl\_health, common\_crawl\_history\_and\_geography, & \\
&  & common\_crawl\_home\_and\_hobbies, common\_crawl\_industrial, & \\
&  & common\_crawl\_literature, common\_crawl\_politics, & \\
&  & common\_crawl\_religion, common\_crawl\_science\_math\_and\_technology, & \\
&  & common\_crawl\_social\_life, common\_crawl\_software, & \\
&  & common\_crawl\_software\_development, common\_crawl\_sports\_and\_fitness, & \\
&  & common\_crawl\_transportation, common\_crawl\_travel\_and\_tourism & \\
\hline
\multirow{3}{*}{OpenLLaMA} & Structural & github\_sample, stackexchange\_sample & 2 \\
& Semi-Structural & arxiv\_sample, wikipedia\_sample & 2 \\
& Free-Text & c4\_sample, common\_crawl\_sample & 2 \\
\hline
\multirow{2}{*}{StarCoder} & \multirow{2}{*}{Structural} & c, c\_sharp, css, dockerfile, go, html, java, javascript, julia, kotlin, & \multirow{2}{*}{21} \\
&  & lua, php, python, r, ruby, rust, scala, shell, sql, swift, typescript &  \\
\hline
\end{tabular}
\caption{Complete domain classification and counts by model family. Domains are categorized as Structural (code/mathematics), Semi-Structural (mixed format), and Free-Text (natural language prose). The \#Dom column shows the count of domains in each category.}
\label{tab:model-domain-classification-complete}
\end{table*}
\subsubsection{Obtaining Token Frequency}
\label{sec: frequency}
In this study, we have used the Infini-gram\footnote{\url{https://infini-gram.readthedocs.io/en/latest/api.html}} to get the token frequency of memorized and unmemorized sequences.
We employ their official Python API to obtain the frequency. 
%However, as the database of Infini-gram is all based on the Llama-2 tokenizer, which is different from the corresponding tokenizer of each model, the frequency is more like an approximate frequency rather than the exact accuracy of each model, as different tokenizers would yield different tokenized results.

We also note that the limitations of our frequency analysis are due to Infini-gram.
First, Infini-gram mainly uses the Llama-2 tokenizer for its n-gram database.
On the contrary, we use the corresponding tokenizer of each model in our other experiments, for example, the Pythia tokenizer for Pythia models.
Furthermore, based on our observations, the Llama-2 tokenizer tends to give shorter tokenized lists compared to the tokenizer of the original corresponding model.
For example, we find that some sequences in the OLMo2 models with 64 tokens are 46 tokens at the Llama-2 tokenizer.
Therefore, we note that our results of frequency analysis are approximate results.
%The Inifini-gram API has a limited access rate for each IP address, where we set 1 minute per query at 4 servers to check the batch frequency of the queried tokens.
%Due to this hard access frequency limit, we only queried the frequency for fully memorized, half-memorized, and unmemorized sequences.

Second, due to the API limit of Infini-gram, we only queried the frequency for fully memorized, half-memorized, and unmemorized sequences.

Third, the Inifini-gram did not provide a Python API endpoint for the OLMo3 data.
%, but provided a web search interface, which does not support queries at a large scale, and the query speed is also slow, so 
This is the reason why we do not report the frequency results for the OLMo3 models. 
%Related results of OLMo3 can be appended if the API access of OLMo3 becomes available.
\subsection{Detailed Distribution Over Domains}
In this section, we show the detailed distribution and exact counts over each domain for all models in the Table \ref{tab:memorized-domain-distribution}.

In this table, we also observed basically the same results, where most of the memorized documents come from codes, the math domain, followed by the arxiv, Wikipedia, and finally free-text domains like PES2O or OCRed documents.
However, some domain classification is also coarse-grained, like Pile-CC in the Pythia and flan in the OLMo2 models, where it may contain documents of various types, making a very accurate distribution specific hard.

In the StarCoder model, since all training data are different codes, we saw a much less polarized distribution regarding the memorized documents distribution compared to the models.
However, we still observed a very organized proportion ordering in different model sizes.
Additionally, this does not seem to strictly relate to the popularity of programming languages, where Python is only at the medium level, and Scala, which is famous for its variable expression, has the largest portion.
Nonetheless, the trend still generally correlates with the popularity and grammaticality of corresponding languages, since Java, C,  HTML, Go, and Kotlin have the highest share, where those languages are either common (C, HTML, Java) or very predictable (Go, Kotlin), and less popular languages (Ruby, R, Julia) have the least proportion.
\subsection{Domain Distribution with Injected Noise}
In this section, we also discuss the distribution over the three domain categories (Structural, Semi-Structural, Free) under different injected relative noise strengths (0.1, 0.3, 0.5) in the Table \ref{tab: memorized_category_proportions_subtables_noise_0p1_0p3_0p5}.

\begin{table*}[t]
\centering
\begin{subtable}{\textwidth}
\centering
\resizebox{\textwidth}{!}{
\begin{tabular}{lccccccccccccc}
\hline
Model & \multicolumn{3}{c}{Pythia} & \multicolumn{2}{c}{OLMo1} & \multicolumn{3}{c}{OLMo2} & \multicolumn{2}{c}{OLMo3} & \multicolumn{3}{c}{OpenLLaMA} \\
\cmidrule(lr){2-4} \cmidrule(lr){5-6} \cmidrule(lr){7-9} \cmidrule(lr){10-11} \cmidrule(lr){12-14}
Size & 1B & 2.8B & 12B & 1B & 7B & \scriptsize{7B} & 13B & 32B & 7B & 32B & 3B & 7B & 13B \\
\hline
Structural & 75.2\% & 76.4\% & 74.4\% & 58.3\% & 51.3\% & 44.9\% & 44.0\% & 39.8\% & 49.0\% & 44.4\% & 66.1\% & 61.5\% & 58.6\% \\
Semi-Structural & 19.6\% & 17.4\% & 17.2\% & 31.1\% & 33.5\% & 9.5\% & 10.5\% & 12.1\% & 6.4\% & 7.9\% & 12.0\% & 13.5\% & 13.1\% \\
Free-Text & 5.1\% & 6.2\% & 8.4\% & 10.6\% & 15.2\% & 45.6\% & 45.5\% & 48.1\% & 44.6\% & 47.7\% & 21.9\% & 25.1\% & 28.3\% \\
\hline
\end{tabular}
}
\caption{Noise = 0.1}
\label{tab: memorized_category_proportions_noise0p1_sub}
\end{subtable}

\vspace{0.1em}

\begin{subtable}{\textwidth}
\centering
\resizebox{\textwidth}{!}{%
\begin{tabular}{lccccccccccccc}
\hline
Model & \multicolumn{3}{c}{Pythia} & \multicolumn{2}{c}{OLMo1} & \multicolumn{3}{c}{OLMo2} & \multicolumn{2}{c}{OLMo3} & \multicolumn{3}{c}{OpenLLaMA} \\
\cmidrule(lr){2-4} \cmidrule(lr){5-6} \cmidrule(lr){7-9} \cmidrule(lr){10-11} \cmidrule(lr){12-14}
Size & 1B & 2.8B & 12B & 1B & 7B & 7B & 13B & 32B & 7B & 32B & 3B & 7B & 13B \\
\hline
Structural & 86.0\% & 80.0\% & 79.6\% & 73.5\% & 86.6\% & 48.2\% & 44.3\% & 40.0\% & 49.8\% & 44.5\% & 67.0\% & 61.6\% & 58.6\% \\
Semi-Structural & 11.4\% & 16.5\% & 15.2\% & 19.6\% & 5.5\% & 10.0\% & 10.5\% & 12.2\% & 6.4\% & 7.8\% & 11.4\% & 13.4\% & 13.1\% \\
Free-Text & 2.6\% & 3.4\% & 5.2\% & 6.9\% & 7.9\% & 41.8\% & 45.2\% & 47.9\% & 43.8\% & 47.7\% & 21.6\% & 25.0\% & 28.3\% \\
\hline
\end{tabular}
}
\caption{Noise = 0.3}
\label{tab:memorized_category_proportions_noise0p3_sub}
\end{subtable}

\vspace{0.1em}

\begin{subtable}{\textwidth}
\centering
\resizebox{\textwidth}{!}{
\begin{tabular}{lccccccccccccc}
\hline
Model & \multicolumn{3}{c}{Pythia} & \multicolumn{2}{c}{OLMo1} & \multicolumn{3}{c}{OLMo2} & \multicolumn{2}{c}{OLMo3} & \multicolumn{3}{c}{OpenLLaMA} \\
\cmidrule(lr){2-4} \cmidrule(lr){5-6} \cmidrule(lr){7-9} \cmidrule(lr){10-11} \cmidrule(lr){12-14}
Size & 1B & 2.8B & 12B & 1B & 7B & 7B & 13B & 32B & 7B & 32B & 3B & 7B & 13B \\
\hline
Structural & 86.6\% & 91.6\% & 91.1\% & 85.3\% & 91.3\% & 55.9\% & 45.2\% & 40.5\% & 54.2\% & 44.6\% & 70.2\% & 61.8\% & 58.6\% \\
Semi-Structural & 8.5\% & 5.9\% & 5.2\% & 7.4\% & 3.3\% & 9.5\% & 10.5\% & 12.3\% & 6.2\% & 7.7\% & 9.1\% & 13.6\% & 13.1\% \\
Free-Text & 4.9\% & 2.5\% & 3.8\% & 7.3\% & 5.4\% & 34.6\% & 44.2\% & 47.2\% & 39.6\% & 47.7\% & 20.7\% & 24.7\% & 28.3\% \\
\hline
\end{tabular}
}
\caption{Noise = 0.5}
\label{tab:memorized_category_proportions_noise0p5_sub}
\end{subtable}

\caption{Proportion of memorized texts (score=1) across Structural, Semi-Structural, and Free-Text categories at different noise levels.}
\label{tab: memorized_category_proportions_subtables_noise_0p1_0p3_0p5}
\end{table*}

From the results, we can see that the distribution of Pythia and OLMo1 is more affected than that of other models.
When increasing the strength of the noise, the distribution largely shifts to the Structural domain, where OLMo1 7b's Structural domain increased from 51.3\% to 91.3\% from changing the noise to 0.5.
Similar results are also observed in the Pythia model, where a 74.4\% increase to 91.1\% at the Structural domain for the 12b model.
However, the same tendency is weaker (OpenLLaMa) or not observable (OLMo2/3) in other models, where the distribution basically does not change.
Especially, the distribution of OLMo3 is basically not affected by the increase in injected noise strength.
The reason may be that more capable models memorize the free texts through semantic information, which is equally robust to the codes or math expressions memorized through repetitive patterns. 

We can observe that injecting stronger noise leads to bottom direction shifting, meaning that the distribution is transferring to the left area, which is expected.
However, we do observe that the OLMo1 models are largely affected by the injected noise, where the highest memorization rate of OLMo1 7b decreased from near 0.1 to 0.001, which is almost 100 times slower than the non-noised model.
Additionally, in most cases, larger models are more robust compared to their small counterparts, where we also observed that the high memorization score probability density of  OLMo2 1b decreases significantly, while larger models like 32b only show very limited influence caused by the noise injection.
Additionally, we also notice that the distribution of those lines is more dispersed at higher noise levels compared to low noise levels.
This means the decreasing trend is not equal for all model families, showing that the robustness to the injected noise is a model-specific feature rather than a common feature that could be shared within a model family.
Such a reason may be due to model parameter differences and training parameters that change with different model sizes.

\subsection{Similarity for MLP Layer}
\begin{figure}[htbp]
    \centering
\includegraphics[width=\linewidth,height=0.5\linewidth]{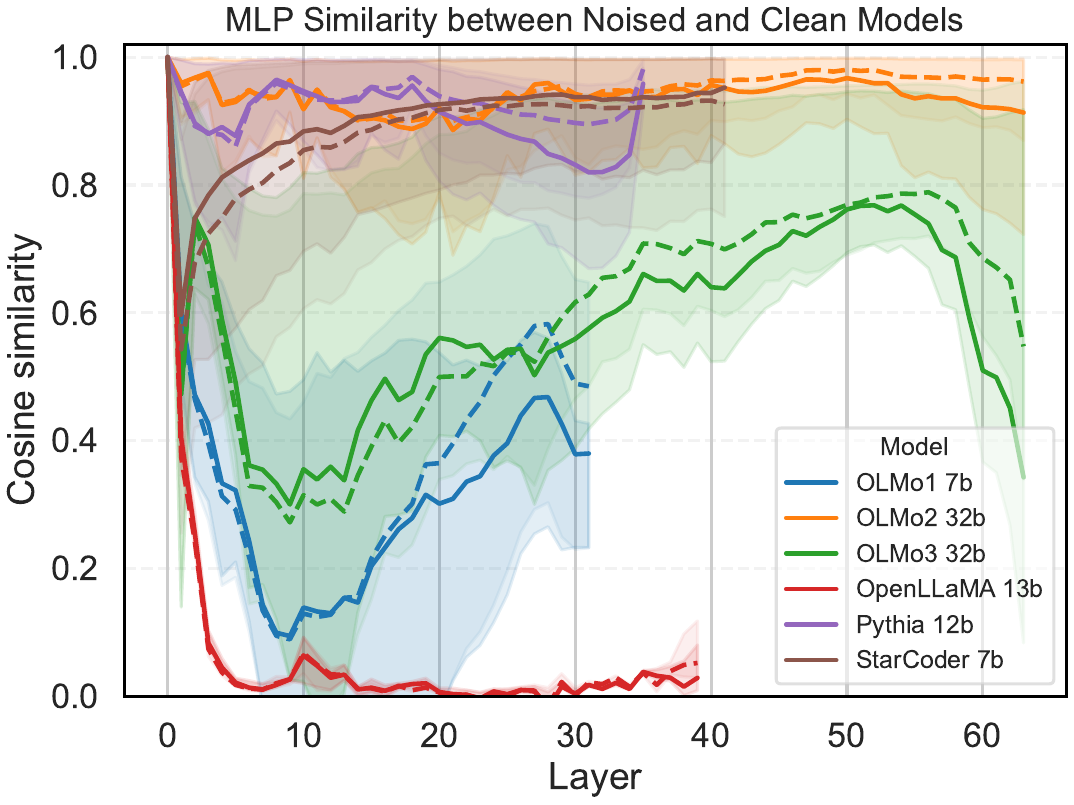}
    \caption{MLP similarity between noised and clean models. Solid/Dotted lines represent the memorized/unmemorized sequence experiment results.}
    \label{fig: MLP component similarity}
\end{figure}

In this section, we show the similarity between noised models and clean models for the MLP layer in the Figure \ref{fig: MLP component similarity}.

In this figure, we saw that the MLP layer shows basically a similar trend to the figure of attention heads.
The similarity also decreases at the beginning layers while bouncing back at later layers.
The similarity of memorized sequences is also lower in the later layers compared to unmemorized sequences.
We also notice that the similarity to the original model is higher at early layers compared to unmemorized sequences.
This means the information itself is strong at early layers, but the decoding requires more specific information for those memorized sequences to be successfully decoded.
It means those memorized sequences provide stronger information clues than those memorized sequences, but decoding them would require very accurate information; in a sense, little perturbed information would drive the LLMs not to generate memorized sequences even though those memorized sequences are highly correlated with model parameters.

Through comparison with the results of the attention head similarity, we also notice that the results are smooth, indicating the internal robustness of each head differs within each layer.
Additionally, we could also observe that the MLP layer tends to have a lower similarity compared to the attention heads.
This may be because the MLP layer aggregates the outputs of all attention heads, so inconsistent similarity across attention heads can also reduce the similarity of the MLP layer to the original model.

\subsection{Examples for Frequency Distribution Including Half-Memorized Sequences}
\begin{figure}[t]
    \centering
\includegraphics[width=\linewidth,height=0.55\linewidth]{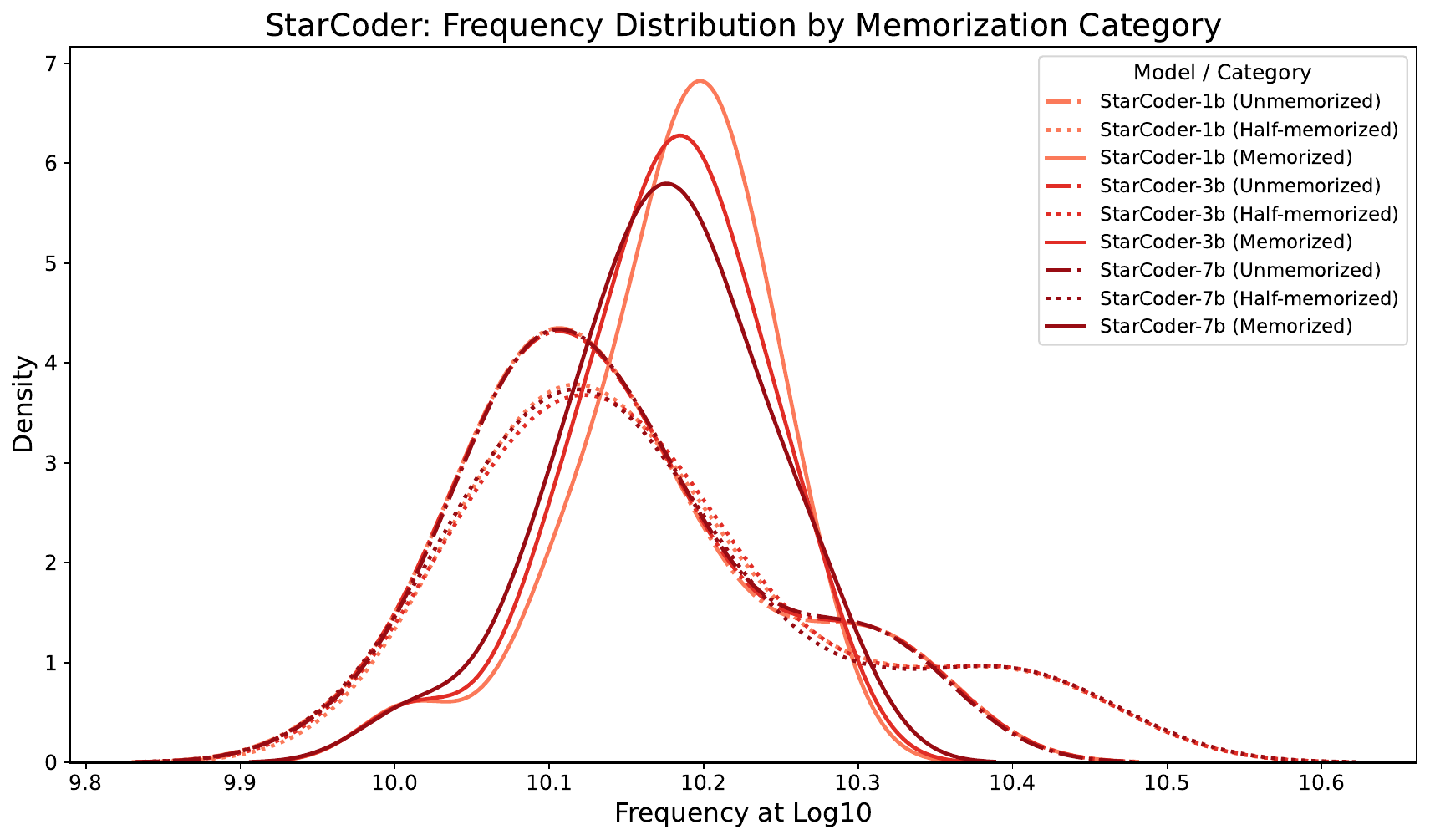}
    \caption{Frequency distribution for memorized, half-memorized, and unmemorized sequences.}
    \label{fig: three way frequency distribution}
\end{figure}

In this section, we present distributions that also include the half-memorized sequences for the StarCoder model family in the Figure \ref{fig: three way frequency distribution}.
\begin{figure*}[t]
\centering
\begin{subfigure}[t]{0.49\textwidth}
    \centering
    \includegraphics[width=\linewidth]{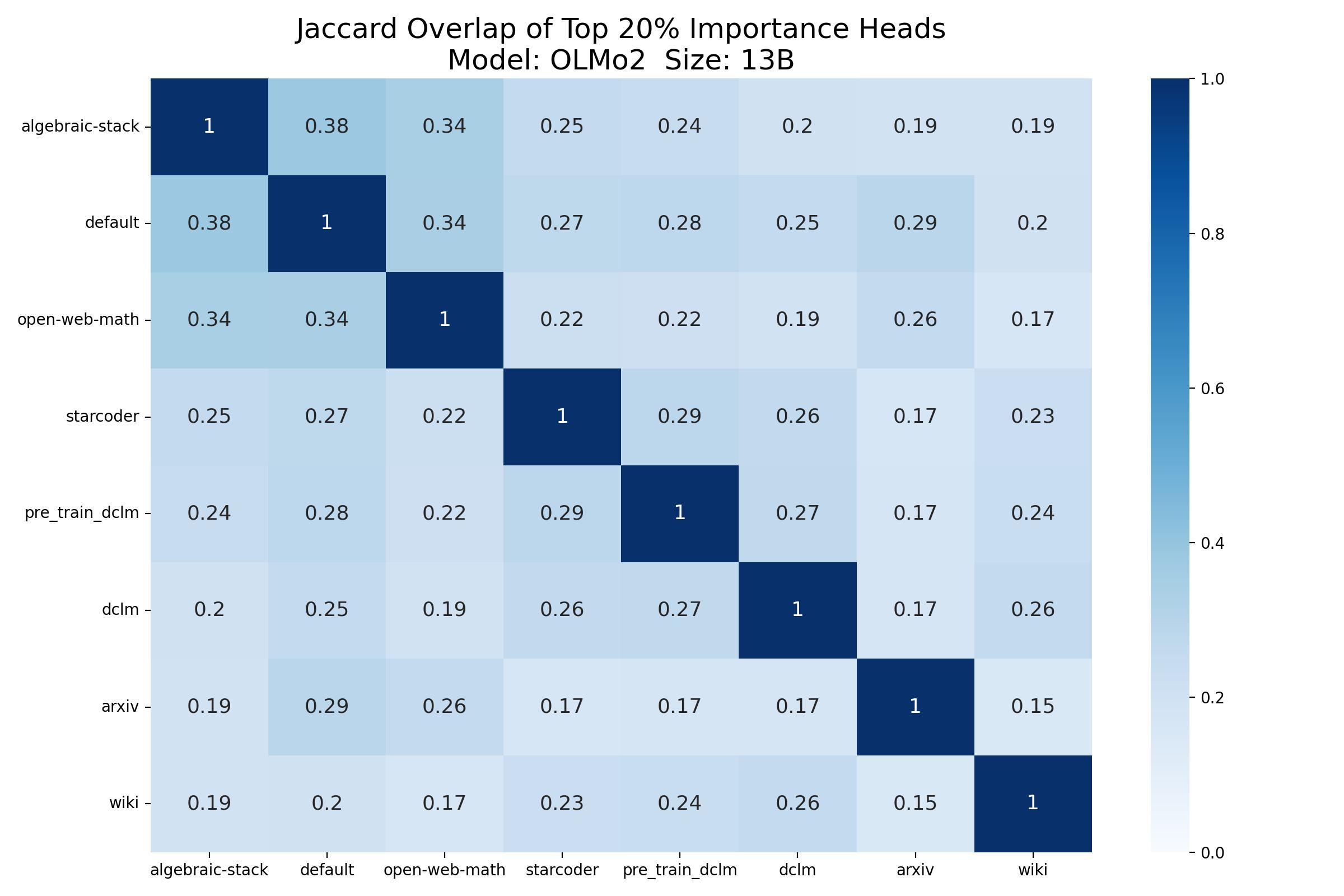}
    \caption{Top 20\% important heads overlap for OLMo2 13b models.}
    \label{fig: olmo2 13b overlap}
\end{subfigure}
\begin{subfigure}[t]{0.49\textwidth}
    \centering
\includegraphics[width=\linewidth,]{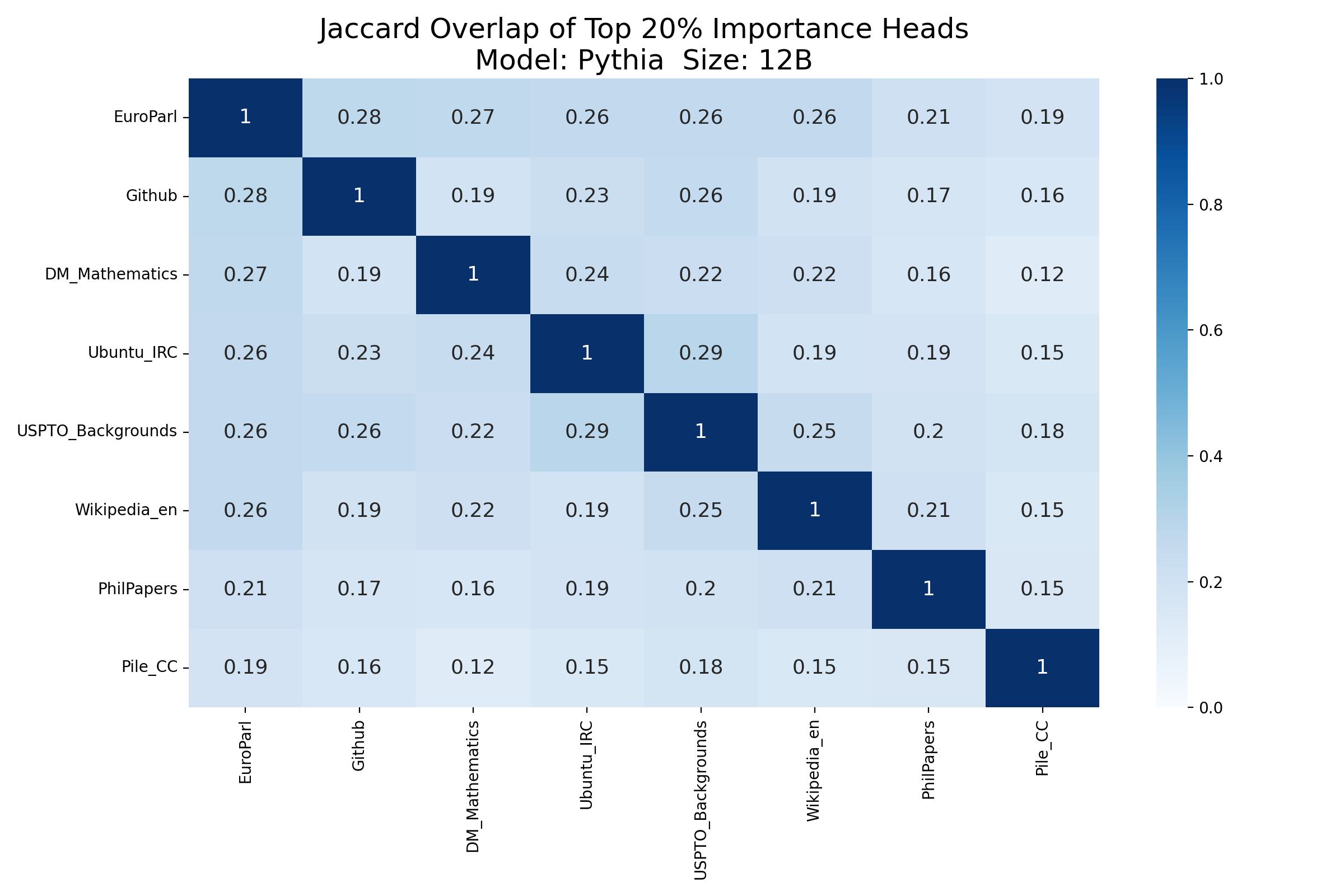}
    \caption{Top 20\% important heads overlap for Pythia 12b models.}
    \label{fig: pythia 12b oevrlap}
\end{subfigure}
\caption{Top 20\%  important heads overlap for Pythia and OLMo2 models on selected domains.}
\label{fig: domain overlap}
\end{figure*}
From these results, we can observe that the frequency of half-memorized sequences is between the fully memorized and unmemorized sequences since the distribution area on the left side is smaller than that of the unmemorized sequences, while larger than that of the memorized sequences.
This indicates, generally, that a higher frequency leads to more memorized sequences.
\subsection{Examples of Shared Import Head Overlap over Domains}

In this section, we show two examples of the overlap ratio between two domains for the Pythia 12b and OLMo2 13b models.
The results are presented in the Figure \ref{fig: domain overlap}.

From this result, we can observe that for any selected domains for those two models. 
They share around 20\% overlap for important attention heads, aligning with the results listed in the Table \ref{tab:head-overlap-random-significance}.
In the left figure, we saw that the algebraic-stack has a higher overlap with open-web-math. 
The reason may be that both domains focus on math-related content.
In the right figure, EuroParl has a higher overlap with other domains in general, while we also notice it is especially close to GitHub and DM Mathematics, reaching around 30\% overlap.
The reason may be that EuroParl contains mainly non-English documents, where those non-English documents are not memorized through semantic understanding but through hard-coded memorization.
This makes the memorization pattern of EuroParl closer to domains that are memorized with hard rules, like codes or math.

\subsection{Averaged Layer Importance of All Models}
\begin{figure*}[tbp] 
\centering % 让图像居中
\includegraphics[width=1\textwidth]{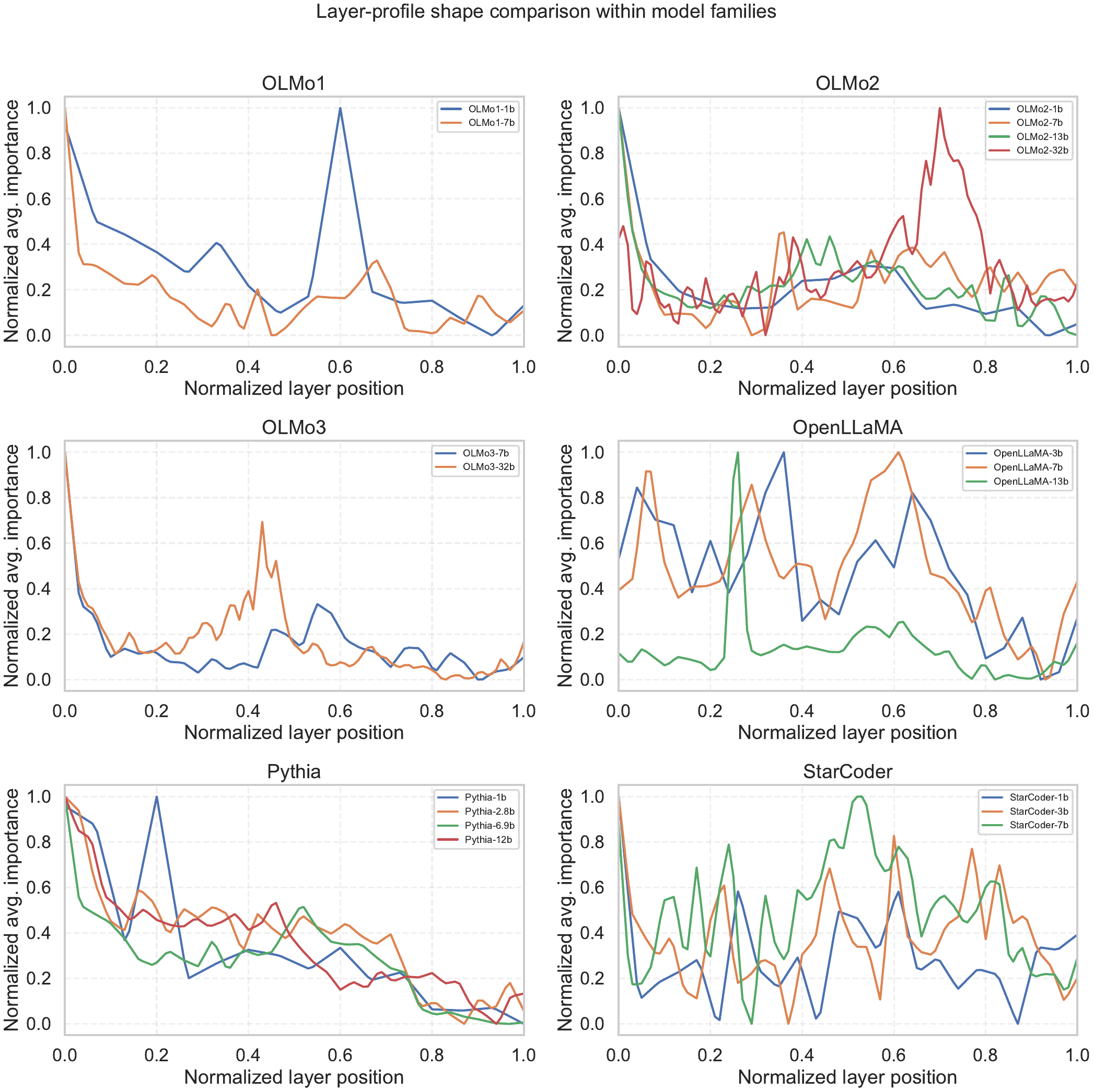}
\caption{Layer-Normalized Averaged Importance Score for Attention Heads Across Layers.}
\label{fig: normalized layer importance}
\end{figure*}

In this section, we present normalized layer importance for different model families in Figure \ref{fig: normalized layer importance} to show the individual layer-level importance structure.

From this feature, we can observe that the normalized distribution in each model family shares certain common features.
For example, for the Pythia models, basically, the layer importance drops across layers for all models from 1b to 12b models.
Additionally, the OLMo1 also presents a similar behavior where the importance score gradually decreases and increases at the middle layers.
Similar patterns are also observed in other model families.
For example, even for the StarCoder, where no similar trends can be observed, all of them show a very unstable layer importance score, no matter the model size.
The reason may be that the model-specific model has a very different internal representation since the learned knowledge is limited to one domain, so the domain-specific knowledge is distributed across the model rather than being constrained in a fixed pattern.
However, when we look at those curves across models, we basically cannot tell their common features, except that beginning layers generally have a high importance score.
\subsection{Examples of Unnormalized Average Layer Importance}
\begin{figure*}[ht]
\centering
\begin{subfigure}[t]{0.49\textwidth}
    \centering
    \includegraphics[width=\linewidth, height=0.55\linewidth]{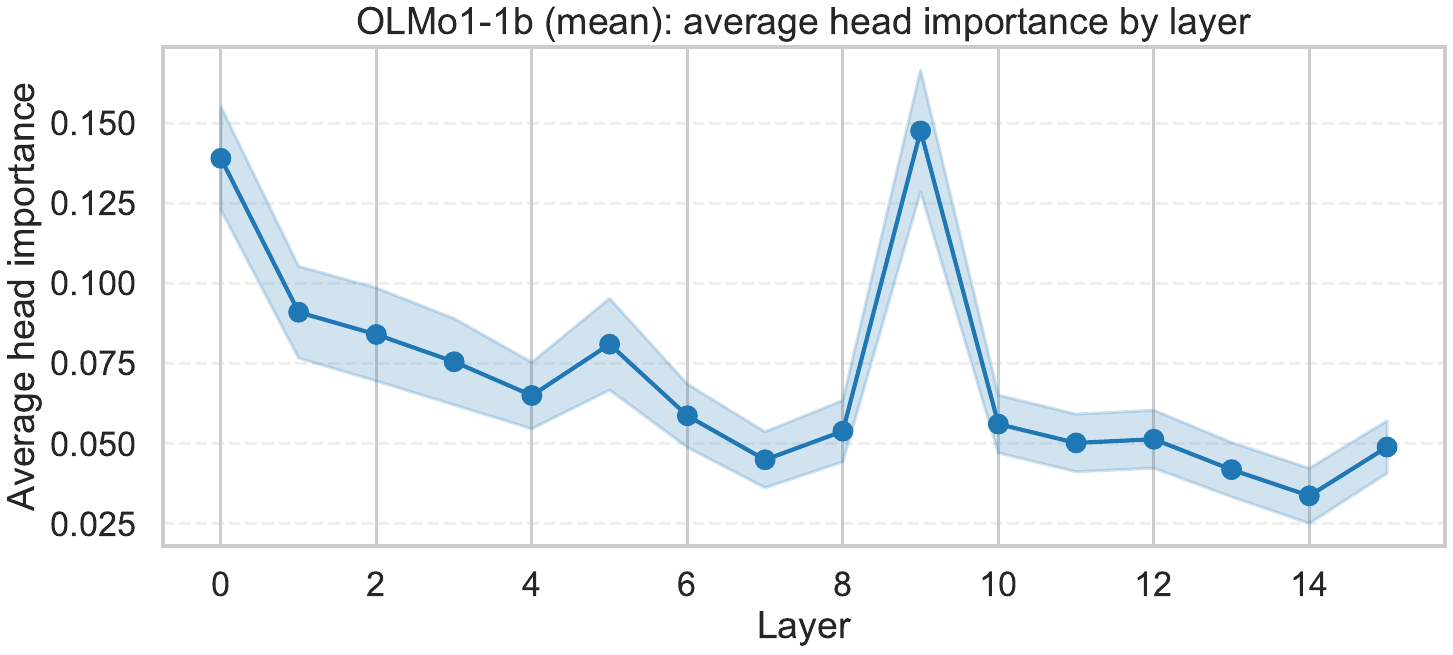}
    \caption{Average layer importance for OLMo1 1b model.}
    \label{fig: layer importance 1b}
\end{subfigure}
\begin{subfigure}[t]{0.49\textwidth}
    \centering
\includegraphics[width=\linewidth,height=0.55\linewidth]{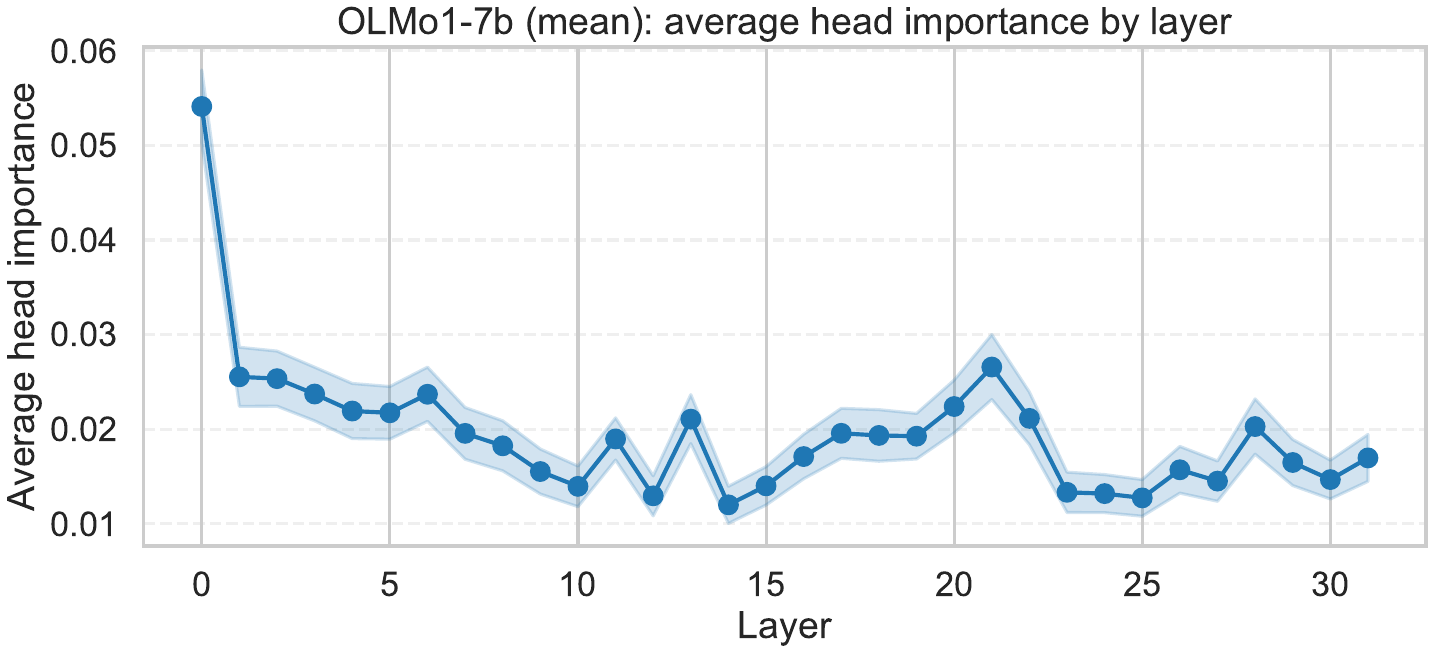}
    \caption{Average layer importance for OLMo1 7b model.}
    \label{fig: layer importance 7b}
\end{subfigure}
\caption{Average layer importance across layers.}
\label{fig: layer importance olmo1}
\end{figure*}
Additionally, we also provide a specific example for the average layer importance of OLMo1 models in Figure \ref{fig: layer importance olmo1}, which is not normalized, serving as an example distribution.
\begin{figure*}[ht]
    \centering
\includegraphics[width=\textwidth, height=0.4\textwidth]{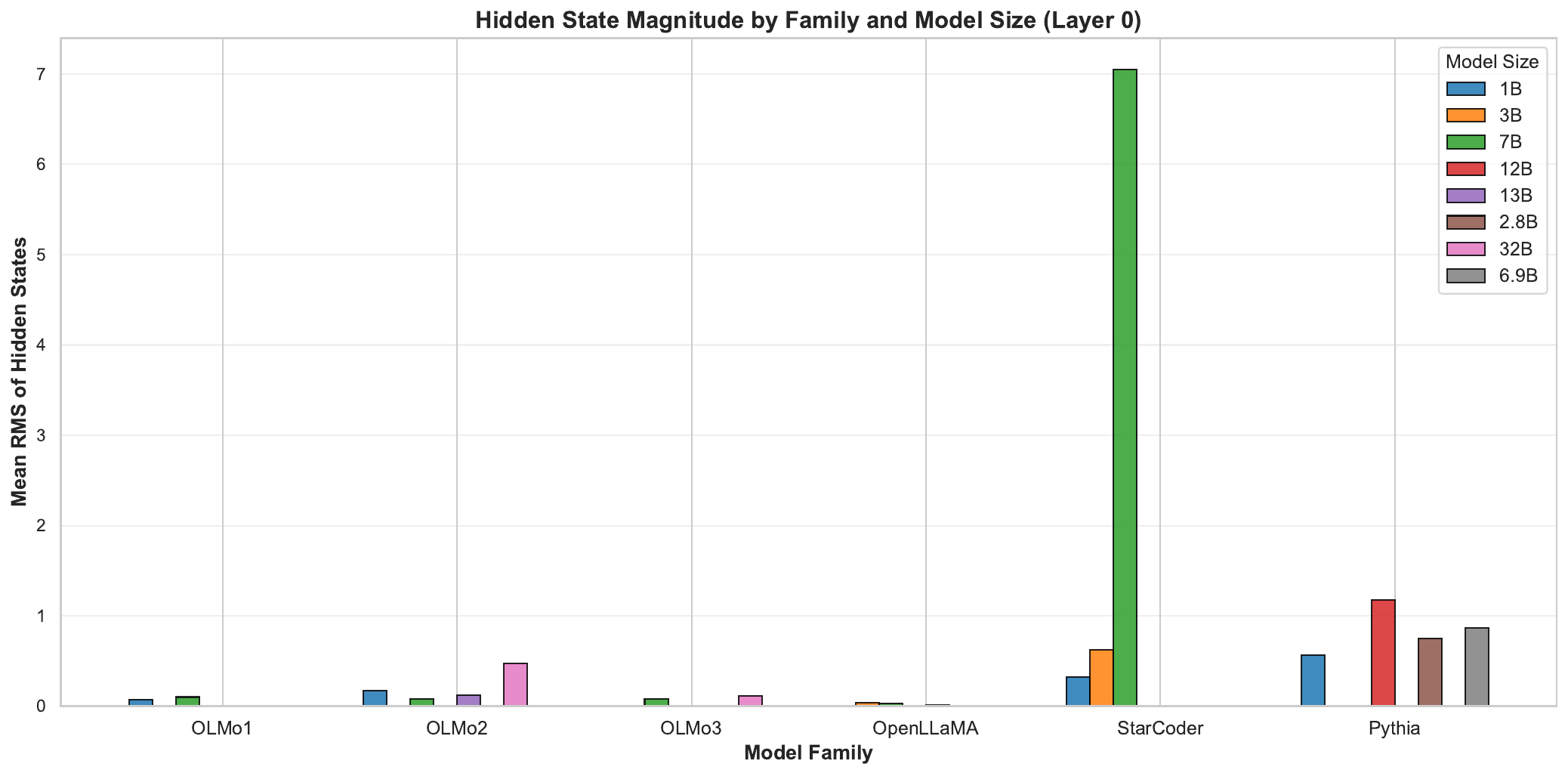}
    \caption{Average norm value of residual stream for different models.}
    \label{fig: rms stream}
\end{figure*}
In this result, we can tell that the average layer importance decreases with the scaling of model size, which is expected since the number of heads also increases with the model size scaling.
At the OLMo1 1b model, removing an attention head at layer 0 gives 0.15 importance score, meaning that 15\% of the output tokens changed compared to the original output.
While at the OLMo1b 7b model, removing one head at the layer 0 gives around 0.05 importance score, indicating only 5\% of output tokens changed compared to the original output.
This change is almost one-third of the 1b model.

\subsection{Absolute Residual Stream Value}

In this study, the injected noise is set relative to the norm value of the residual stream.
However, as different models have different MLP structures and training processes,  the exact norm value of each model also differs across model families.
In this section, we present the norm value of the residual stream at the first layer as a reference material for the relative noise injection.
The results are presented in the Figure \ref{fig: rms stream}.

From the results, we can see that the residual stream of different models differs a lot.
The StarCoder and Pythia families have a higher residual stream value, especially the StarCoder 7b, which shows a significant burst at the 7b models.
Their relatively larger value also leads to a large absolute injected noise, explaining why they are less robust to noise perturbation.
This could also be seen from the OpenLLaMa model, which has the lowest average residual stream norm.
This also makes this model family most robust to noise, reflected in their memorization distribution shift.
OLMo1/2/3 families show a similar value range, showing the effect brought by the inherited training recipe.

\clearpage
\clearpage

\clearpage
\begin{figure*}[htbp] 
\centering
\includegraphics[width=\textwidth]{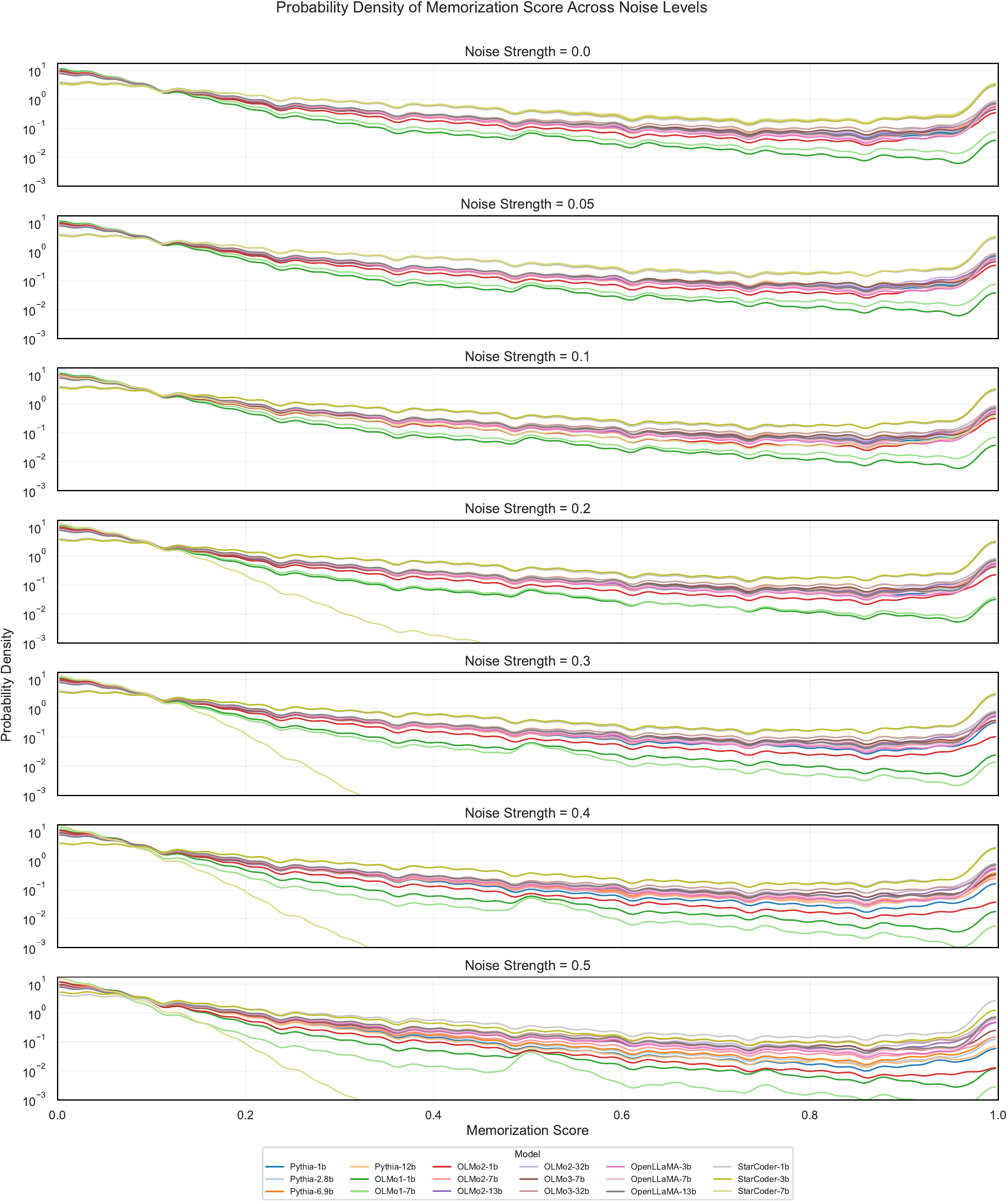}
\caption{Decoding probability (Y-axis) for memorized and unmemorized tokens across different layers (X-axis) for all models. The solid/dotted line represents the decoding probability across layers for memorized/unmemorized sequences.}
\label{fig: memorization score distribution noise}
\end{figure*}
\clearpage

\clearpage
\onecolumn
\begin{longtable}{l l l r c r}
\toprule
\textbf{Model} & \textbf{Size} & \textbf{Domain} & \textbf{Count} & \textbf{Share} & \textbf{Total} \\
\midrule
\endfirsthead
\toprule
\textbf{Model} & \textbf{Size} & \textbf{Domain} & \textbf{Count} & \textbf{Share} & \textbf{Total} \\
\midrule
\endhead
\multicolumn{6}{r}{\textit{Continued on next page}} \\
\endfoot
\endlastfoot
\multirow{38}{*}{OLMo1} & \multirow{19}{*}{1B} & starcoder & 2,834 & 48.99\% & \multirow{19}{*}{5,785} \\
 &  & cc\_en\_tail & 575 & 9.94\% &  \\
 &  & c4\_filtered & 367 & 6.34\% &  \\
 &  & books & 334 & 5.77\% &  \\
 &  & cc\_news\_head & 318 & 5.50\% &  \\
 &  & proof\_pile\_2\_open\_web\_math & 295 & 5.10\% &  \\
 &  & cc\_news\_tail & 202 & 3.49\% &  \\
 &  & cc\_en\_middle & 185 & 3.20\% &  \\
 &  & proof\_pile\_2\_algebraic\_stack & 175 & 3.03\% &  \\
 &  & falcon\_refinedweb\_filtered & 118 & 2.04\% &  \\
 &  & cc\_news\_middle & 113 & 1.95\% &  \\
 &  & cc\_en\_head & 97 & 1.68\% &  \\
 &  & wikiref\_megawika & 53 & 0.92\% &  \\
 &  & wiki & 42 & 0.73\% &  \\
 &  & redpajama\_arxiv & 34 & 0.59\% &  \\
 &  & reddit & 27 & 0.47\% &  \\
 &  & pes2o & 9 & 0.16\% &  \\
 &  & tulu\_flan & 4 & 0.07\% &  \\
 &  & redpajama\_stackexchange & 3 & 0.05\% &  \\
\cmidrule{2-6}
 & \multirow{19}{*}{7B} & starcoder & 3,952 & 35.01\% & \multirow{19}{*}{11,287} \\
 &  & cc\_en\_tail & 994 & 8.81\% &  \\
 &  & proof\_pile\_2\_open\_web\_math & 981 & 8.69\% &  \\
 &  & books & 940 & 8.33\% &  \\
 &  & c4\_filtered & 830 & 7.35\% &  \\
 &  & proof\_pile\_2\_algebraic\_stack & 679 & 6.02\% &  \\
 &  & cc\_news\_tail & 613 & 5.43\% &  \\
 &  & cc\_news\_head & 530 & 4.70\% &  \\
 &  & cc\_en\_middle & 342 & 3.03\% &  \\
 &  & cc\_news\_middle & 310 & 2.75\% &  \\
 &  & falcon\_refinedweb\_filtered & 215 & 1.90\% &  \\
 &  & wiki & 207 & 1.83\% &  \\
 &  & cc\_en\_head & 196 & 1.74\% &  \\
 &  & wikiref\_megawika & 150 & 1.33\% &  \\
 &  & redpajama\_arxiv & 92 & 0.82\% &  \\
 &  & tulu\_flan & 92 & 0.82\% &  \\
 &  & reddit & 72 & 0.64\% &  \\
 &  & pes2o & 66 & 0.58\% &  \\
 &  & redpajama\_stackexchange & 26 & 0.23\% &  \\
\midrule
 & \multirow{14}{*}{13B} & flan & 17,133 & 24.80\% & \multirow{14}{*}{69,080} \\
 &  & starcoder & 14,447 & 20.91\% &  \\
 &  & math & 10,152 & 14.70\% &  \\
 &  & algebraic-stack & 7,072 & 10.24\% &  \\
 &  & open-web-math & 5,439 & 7.87\% &  \\
 &  & arxiv & 4,000 & 5.79\% &  \\
 &  & dclm & 3,251 & 4.71\% &  \\
 &  & default & 2,679 & 3.88\% &  \\
 &  & pre\_train\_dclm & 1,804 & 2.61\% &  \\
 &  & wiki & 1,377 & 1.99\% &  \\
 &  & pre\_train\_wiki & 1,293 & 1.87\% &  \\
 &  & stackexchange & 280 & 0.41\% &  \\
 &  & pre\_train\_pes2o & 81 & 0.12\% &  \\
 &  & pes2o & 72 & 0.10\% &  \\
\cmidrule{2-6}
 & \multirow{14}{*}{1B} & flan & 12,291 & 31.34\% & \multirow{14}{*}{39,219} \\
 &  & starcoder & 8,188 & 20.88\% &  \\
 &  & math & 5,223 & 13.32\% &  \\
 &  & algebraic-stack & 3,868 & 9.86\% &  \\
 &  & open-web-math & 3,766 & 9.60\% &  \\
 &  & arxiv & 2,457 & 6.26\% &  \\
 &  & dclm & 1,217 & 3.10\% &  \\
 &  & default & 1,169 & 2.98\% &  \\
 &  & pre\_train\_dclm & 600 & 1.53\% &  \\
 &  & pre\_train\_wiki & 184 & 0.47\% &  \\
 &  & wiki & 159 & 0.41\% &  \\
 &  & stackexchange & 83 & 0.21\% &  \\
 &  & pes2o & 10 & 0.03\% &  \\
 &  & pre\_train\_pes2o & 4 & 0.01\% &  \\
\cmidrule{2-6}
 & \multirow{14}{*}{32B} & flan & 19,828 & 20.59\% & \multirow{14}{*}{96,284} \\
 &  & starcoder & 19,185 & 19.93\% &  \\
 &  & math & 11,390 & 11.83\% &  \\
 &  & algebraic-stack & 10,433 & 10.84\% &  \\
 &  & dclm & 8,958 & 9.30\% &  \\
 &  & open-web-math & 7,130 & 7.41\% &  \\
 &  & arxiv & 5,389 & 5.60\% &  \\
 &  & default & 4,527 & 4.70\% &  \\
 &  & pre\_train\_dclm & 3,624 & 3.76\% &  \\
 &  & wiki & 2,593 & 2.69\% &  \\
 &  & pre\_train\_wiki & 2,451 & 2.55\% &  \\
 &  & stackexchange & 589 & 0.61\% &  \\
 &  & pre\_train\_pes2o & 95 & 0.10\% &  \\
 &  & pes2o & 92 & 0.10\% &  \\
\cmidrule{2-6}
 & \multirow{14}{*}{7B} & flan & 14,803 & 24.46\% & \multirow{14}{*}{60,509} \\
 &  & starcoder & 13,729 & 22.69\% &  \\
 &  & math & 8,142 & 13.46\% &  \\
 &  & algebraic-stack & 7,515 & 12.42\% &  \\
 &  & open-web-math & 5,035 & 8.32\% &  \\
 &  & arxiv & 3,603 & 5.95\% &  \\
 &  & default & 2,297 & 3.80\% &  \\
 &  & dclm & 2,256 & 3.73\% &  \\
 &  & pre\_train\_dclm & 1,436 & 2.37\% &  \\
 &  & pre\_train\_wiki & 774 & 1.28\% &  \\
 &  & wiki & 727 & 1.20\% &  \\
 &  & stackexchange & 161 & 0.27\% &  \\
 &  & pre\_train\_pes2o & 17 & 0.03\% &  \\
 &  & pes2o & 14 & 0.02\% &  \\
\midrule
 & \multirow{47}{*}{32B} & stack\_edu\_Java & 12,600 & 5.15\% & \multirow{47}{*}{244,693} \\
 &  & stack\_edu\_PHP & 10,714 & 4.38\% &  \\
 &  & stack\_edu\_SQL & 10,488 & 4.29\% &  \\
 &  & common\_crawl\_history\_and\_geography & 9,791 & 4.00\% &  \\
 &  & stack\_edu\_Go & 9,535 & 3.90\% &  \\
 &  & finemath\_3plus & 8,732 & 3.57\% &  \\
 &  & stack\_edu\_Rust & 8,363 & 3.42\% &  \\
 &  & stack\_edu\_Ruby & 8,066 & 3.30\% &  \\
 &  & stack\_edu\_Python & 7,392 & 3.02\% &  \\
 &  & common\_crawl\_software & 6,892 & 2.82\% &  \\
 &  & common\_crawl\_software\_development & 6,652 & 2.72\% &  \\
 &  & common\_crawl\_travel\_and\_tourism & 6,603 & 2.70\% &  \\
 &  & rpj\_proofpile\_arxiv & 6,478 & 2.65\% &  \\
 &  & common\_crawl\_transportation & 6,412 & 2.62\% &  \\
 &  & common\_crawl\_religion & 6,257 & 2.56\% &  \\
 &  & stack\_edu\_JavaScript & 6,075 & 2.48\% &  \\
 &  & common\_crawl\_literature & 6,071 & 2.48\% &  \\
 &  & stack\_edu\_Cpp & 5,986 & 2.45\% &  \\
 &  & stack\_edu\_C & 5,753 & 2.35\% &  \\
 &  & common\_crawl\_science\_math\_and\_technology & 5,729 & 2.34\% &  \\
 &  & stack\_edu\_Swift & 5,701 & 2.33\% &  \\
 &  & common\_crawl\_home\_and\_hobbies & 5,689 & 2.32\% &  \\
 &  & common\_crawl\_games & 5,676 & 2.32\% &  \\
 &  & common\_crawl\_finance\_and\_business & 5,569 & 2.28\% &  \\
 &  & common\_crawl\_electronics\_and\_hardware & 5,467 & 2.23\% &  \\
 &  & common\_crawl\_entertainment & 5,317 & 2.17\% &  \\
 &  & stack\_edu\_Shell & 5,303 & 2.17\% &  \\
 &  & stack\_edu\_Markdown & 5,258 & 2.15\% &  \\
 &  & common\_crawl\_industrial & 4,311 & 1.76\% &  \\
 &  & common\_crawl\_crime\_and\_law & 4,099 & 1.68\% &  \\
 &  & common\_crawl\_education\_and\_jobs & 4,000 & 1.63\% &  \\
 &  & common\_crawl\_health & 3,853 & 1.57\% &  \\
 &  & stack\_edu\_CSharp & 3,747 & 1.53\% &  \\
 &  & stack\_edu\_TypeScript & 3,652 & 1.49\% &  \\
 &  & common\_crawl\_social\_life & 3,159 & 1.29\% &  \\
 &  & common\_crawl\_food\_and\_dining & 3,068 & 1.25\% &  \\
 &  & common\_crawl\_art\_and\_design & 3,043 & 1.24\% &  \\
 &  & common\_crawl\_adult\_content & 2,932 & 1.20\% &  \\
 &  & common\_crawl\_politics & 2,806 & 1.15\% &  \\
 &  & common\_crawl\_sports\_and\_fitness & 1,865 & 0.76\% &  \\
 &  & dolma1\_7\_wiki\_en & 1,802 & 0.74\% &  \\
 &  & olmocr\_science\_pdfs\_education\_and\_jobs & 1,711 & 0.70\% &  \\
 &  & common\_crawl\_fashion\_and\_beauty & 1,506 & 0.62\% &  \\
 &  & olmocr\_science\_pdfs\_art\_and\_design & 282 & 0.12\% &  \\
 &  & olmocr\_science\_pdfs\_entertainment & 267 & 0.11\% &  \\
 &  & olmocr\_science\_pdfs\_sports\_and\_fitness & 13 & 0.01\% &  \\
 &  & olmocr\_science\_pdfs\_adult\_content & 8 & 0.00\% &  \\
\cmidrule{2-6}
 & \multirow{47}{*}{7B} & stack\_edu\_Java & 9,975 & 6.07\% & \multirow{47}{*}{164,337} \\
 &  & stack\_edu\_SQL & 8,621 & 5.25\% &  \\
 &  & stack\_edu\_Go & 7,712 & 4.69\% &  \\
 &  & stack\_edu\_PHP & 7,434 & 4.52\% &  \\
 &  & common\_crawl\_history\_and\_geography & 6,591 & 4.01\% &  \\
 &  & stack\_edu\_Rust & 6,484 & 3.95\% &  \\
 &  & stack\_edu\_Ruby & 5,625 & 3.42\% &  \\
 &  & stack\_edu\_Python & 5,147 & 3.13\% &  \\
 &  & finemath\_3plus & 4,764 & 2.90\% &  \\
 &  & common\_crawl\_software\_development & 4,675 & 2.84\% &  \\
 &  & stack\_edu\_C & 4,493 & 2.73\% &  \\
 &  & rpj\_proofpile\_arxiv & 4,239 & 2.58\% &  \\
 &  & stack\_edu\_JavaScript & 4,209 & 2.56\% &  \\
 &  & stack\_edu\_Shell & 4,193 & 2.55\% &  \\
 &  & common\_crawl\_software & 4,192 & 2.55\% &  \\
 &  & stack\_edu\_Swift & 4,177 & 2.54\% &  \\
 &  & stack\_edu\_Cpp & 4,112 & 2.50\% &  \\
 &  & common\_crawl\_travel\_and\_tourism & 4,080 & 2.48\% &  \\
 &  & common\_crawl\_literature & 4,038 & 2.46\% &  \\
 &  & common\_crawl\_games & 3,851 & 2.34\% &  \\
 &  & common\_crawl\_home\_and\_hobbies & 3,764 & 2.29\% &  \\
 &  & common\_crawl\_entertainment & 3,696 & 2.25\% &  \\
 &  & common\_crawl\_science\_math\_and\_technology & 3,640 & 2.21\% &  \\
 &  & common\_crawl\_transportation & 3,525 & 2.14\% &  \\
 &  & common\_crawl\_electronics\_and\_hardware & 3,490 & 2.12\% &  \\
 &  & common\_crawl\_religion & 3,343 & 2.03\% &  \\
 &  & common\_crawl\_finance\_and\_business & 3,243 & 1.97\% &  \\
 &  & stack\_edu\_Markdown & 3,087 & 1.88\% &  \\
 &  & common\_crawl\_crime\_and\_law & 3,040 & 1.85\% &  \\
 &  & common\_crawl\_industrial & 2,820 & 1.72\% &  \\
 &  & stack\_edu\_CSharp & 2,655 & 1.62\% &  \\
 &  & stack\_edu\_TypeScript & 2,586 & 1.57\% &  \\
 &  & common\_crawl\_education\_and\_jobs & 2,527 & 1.54\% &  \\
 &  & common\_crawl\_health & 2,065 & 1.26\% &  \\
 &  & common\_crawl\_politics & 1,927 & 1.17\% &  \\
 &  & common\_crawl\_art\_and\_design & 1,885 & 1.15\% &  \\
 &  & common\_crawl\_social\_life & 1,807 & 1.10\% &  \\
 &  & common\_crawl\_adult\_content & 1,721 & 1.05\% &  \\
 &  & common\_crawl\_food\_and\_dining & 1,630 & 0.99\% &  \\
 &  & common\_crawl\_sports\_and\_fitness & 1,040 & 0.63\% &  \\
 &  & common\_crawl\_fashion\_and\_beauty & 788 & 0.48\% &  \\
 &  & olmocr\_science\_pdfs\_education\_and\_jobs & 672 & 0.41\% &  \\
 &  & dolma1\_7\_wiki\_en & 396 & 0.24\% &  \\
 &  & olmocr\_science\_pdfs\_art\_and\_design & 214 & 0.13\% &  \\
 &  & olmocr\_science\_pdfs\_entertainment & 154 & 0.09\% &  \\
 &  & olmocr\_science\_pdfs\_adult\_content & 6 & 0.00\% &  \\
 &  & olmocr\_science\_pdfs\_sports\_and\_fitness & 4 & 0.00\% &  \\
\midrule
 & \multirow{6}{*}{13B} & github\_sample & 18,335 & 58.46\% & \multirow{6}{*}{31,365} \\
 &  & common\_crawl\_sample & 5,687 & 18.13\% &  \\
 &  & wikipedia\_sample & 3,194 & 10.18\% &  \\
 &  & c4\_sample & 3,137 & 10.00\% &  \\
 &  & arxiv\_sample & 936 & 2.98\% &  \\
 &  & stackexchange\_sample & 76 & 0.24\% &  \\
\cmidrule{2-6}
 & \multirow{6}{*}{3B} & github\_sample & 15,029 & 65.76\% & \multirow{6}{*}{22,855} \\
 &  & common\_crawl\_sample & 2,797 & 12.24\% &  \\
 &  & c4\_sample & 2,211 & 9.67\% &  \\
 &  & wikipedia\_sample & 2,121 & 9.28\% &  \\
 &  & arxiv\_sample & 654 & 2.86\% &  \\
 &  & stackexchange\_sample & 43 & 0.19\% &  \\
\cmidrule{2-6}
 & \multirow{6}{*}{7B} & github\_sample & 16,290 & 61.05\% & \multirow{6}{*}{26,681} \\
 &  & common\_crawl\_sample & 3,982 & 14.92\% &  \\
 &  & wikipedia\_sample & 2,877 & 10.78\% &  \\
 &  & c4\_sample & 2,749 & 10.30\% &  \\
 &  & arxiv\_sample & 726 & 2.72\% &  \\
 &  & stackexchange\_sample & 57 & 0.21\% &  \\
\midrule
 & \multirow{16}{*}{12B} & Github & 8,380 & 78.71\% & \multirow{16}{*}{10,647} \\
 &  & Pile-CC & 1,427 & 13.40\% &  \\
 &  & FreeLaw & 420 & 3.94\% &  \\
 &  & USPTO Backgrounds & 134 & 1.26\% &  \\
 &  & PubMed Central & 127 & 1.19\% &  \\
 &  & NIH ExPorter & 47 & 0.44\% &  \\
 &  & Wikipedia (en) & 40 & 0.38\% &  \\
 &  & Enron Emails & 27 & 0.25\% &  \\
 &  & StackExchange & 27 & 0.25\% &  \\
 &  & ArXiv & 6 & 0.06\% &  \\
 &  & HackerNews & 4 & 0.04\% &  \\
 &  & Gutenberg (PG-19) & 3 & 0.03\% &  \\
 &  & EuroParl & 2 & 0.02\% &  \\
 &  & PhilPapers & 1 & 0.01\% &  \\
 &  & PubMed Abstracts & 1 & 0.01\% &  \\
 &  & Ubuntu IRC & 1 & 0.01\% &  \\
\cmidrule{2-6}
 & \multirow{6}{*}{160M} & Github & 2,082 & 91.40\% & \multirow{6}{*}{2,278} \\
 &  & FreeLaw & 107 & 4.70\% &  \\
 &  & Pile-CC & 51 & 2.24\% &  \\
 &  & PubMed Central & 36 & 1.58\% &  \\
 &  & PhilPapers & 1 & 0.04\% &  \\
 &  & StackExchange & 1 & 0.04\% &  \\
\cmidrule{2-6}
 & \multirow{13}{*}{1B} & Github & 6,929 & 82.80\% & \multirow{13}{*}{8,368} \\
 &  & Pile-CC & 815 & 9.74\% &  \\
 &  & FreeLaw & 353 & 4.22\% &  \\
 &  & PubMed Central & 125 & 1.49\% &  \\
 &  & USPTO Backgrounds & 60 & 0.72\% &  \\
 &  & NIH ExPorter & 45 & 0.54\% &  \\
 &  & StackExchange & 14 & 0.17\% &  \\
 &  & Wikipedia (en) & 10 & 0.12\% &  \\
 &  & Enron Emails & 7 & 0.08\% &  \\
 &  & ArXiv & 3 & 0.04\% &  \\
 &  & Gutenberg (PG-19) & 3 & 0.04\% &  \\
 &  & HackerNews & 3 & 0.04\% &  \\
 &  & PhilPapers & 1 & 0.01\% &  \\
\cmidrule{2-6}
 & \multirow{14}{*}{2.8B} & Github & 7,575 & 82.81\% & \multirow{14}{*}{9,147} \\
 &  & Pile-CC & 805 & 8.80\% &  \\
 &  & FreeLaw & 369 & 4.03\% &  \\
 &  & USPTO Backgrounds & 141 & 1.54\% &  \\
 &  & PubMed Central & 126 & 1.38\% &  \\
 &  & NIH ExPorter & 45 & 0.49\% &  \\
 &  & Wikipedia (en) & 29 & 0.32\% &  \\
 &  & Enron Emails & 26 & 0.28\% &  \\
 &  & StackExchange & 18 & 0.20\% &  \\
 &  & ArXiv & 5 & 0.05\% &  \\
 &  & HackerNews & 4 & 0.04\% &  \\
 &  & PubMed Abstracts & 2 & 0.02\% &  \\
 &  & Gutenberg (PG-19) & 1 & 0.01\% &  \\
 &  & PhilPapers & 1 & 0.01\% &  \\
\cmidrule{2-6}
 & \multirow{12}{*}{410M} & Github & 4,995 & 85.24\% & \multirow{12}{*}{5,860} \\
 &  & Pile-CC & 411 & 7.01\% &  \\
 &  & FreeLaw & 254 & 4.33\% &  \\
 &  & PubMed Central & 116 & 1.98\% &  \\
 &  & NIH ExPorter & 45 & 0.77\% &  \\
 &  & USPTO Backgrounds & 25 & 0.43\% &  \\
 &  & StackExchange & 9 & 0.15\% &  \\
 &  & ArXiv & 1 & 0.02\% &  \\
 &  & Enron Emails & 1 & 0.02\% &  \\
 &  & Gutenberg (PG-19) & 1 & 0.02\% &  \\
 &  & PhilPapers & 1 & 0.02\% &  \\
 &  & Wikipedia (en) & 1 & 0.02\% &  \\
\cmidrule{2-6}
 & \multirow{15}{*}{6.9B} & Github & 8,350 & 79.29\% & \multirow{15}{*}{10,531} \\
 &  & Pile-CC & 1,376 & 13.07\% &  \\
 &  & FreeLaw & 405 & 3.85\% &  \\
 &  & PubMed Central & 128 & 1.22\% &  \\
 &  & USPTO Backgrounds & 127 & 1.21\% &  \\
 &  & NIH ExPorter & 46 & 0.44\% &  \\
 &  & Enron Emails & 27 & 0.26\% &  \\
 &  & Wikipedia (en) & 27 & 0.26\% &  \\
 &  & StackExchange & 23 & 0.22\% &  \\
 &  & ArXiv & 7 & 0.07\% &  \\
 &  & HackerNews & 7 & 0.07\% &  \\
 &  & Gutenberg (PG-19) & 4 & 0.04\% &  \\
 &  & PubMed Abstracts & 2 & 0.02\% &  \\
 &  & PhilPapers & 1 & 0.01\% &  \\
 &  & Ubuntu IRC & 1 & 0.01\% &  \\
\midrule
 & \multirow{11}{*}{1B} & scala & 69,591 & 16.30\% & \multirow{11}{*}{426,831} \\
 &  & java & 59,360 & 13.91\% &  \\
 &  & html & 43,000 & 10.07\% &  \\
 &  & c & 42,895 & 10.05\% &  \\
 &  & go & 38,165 & 8.94\% &  \\
 &  & kotlin & 28,411 & 6.66\% &  \\
 &  & c\_sharp & 18,946 & 4.44\% &  \\
 &  & php & 18,309 & 4.29\% &  \\
 &  & rust & 17,913 & 4.20\% &  \\
 &  & python & 16,425 & 3.85\% &  \\
 &  & javascript & 11,033 & 2.58\% &  \\
 &  & swift & 9,282 & 2.17\% &  \\
 &  & shell & 8,943 & 2.10\% &  \\
 &  & typescript & 8,793 & 2.06\% &  \\
 &  & sql & 7,888 & 1.85\% &  \\
 &  & ruby & 7,659 & 1.79\% &  \\
 &  & lua & 7,534 & 1.77\% &  \\
 &  & css & 5,822 & 1.36\% &  \\
 &  & dockerfile & 3,779 & 0.89\% &  \\
 &  & julia & 2,811 & 0.66\% &  \\
 &  & r & 272 & 0.06\% &  \\
\cmidrule{2-6}
 & \multirow{21}{*}{3B} & scala & 72,282 & 15.24\% & \multirow{21}{*}{474,193} \\
 &  & java & 63,976 & 13.49\% &  \\
 &  & c & 48,425 & 10.21\% &  \\
 &  & html & 45,264 & 9.55\% &  \\
 &  & go & 41,146 & 8.68\% &  \\
 &  & kotlin & 31,384 & 6.62\% &  \\
 &  & c\_sharp & 22,748 & 4.80\% &  \\
 &  & php & 21,139 & 4.46\% &  \\
 &  & rust & 20,112 & 4.24\% &  \\
 &  & python & 18,807 & 3.97\% &  \\
 &  & javascript & 12,895 & 2.72\% &  \\
 &  & lua & 10,242 & 2.16\% &  \\
 &  & shell & 10,195 & 2.15\% &  \\
 &  & sql & 10,172 & 2.15\% &  \\
 &  & swift & 10,031 & 2.12\% &  \\
 &  & typescript & 9,652 & 2.04\% &  \\
 &  & ruby & 9,273 & 1.96\% &  \\
 &  & css & 7,262 & 1.53\% &  \\
 &  & dockerfile & 4,688 & 0.99\% &  \\
 &  & julia & 3,789 & 0.80\% &  \\
 &  & r & 711 & 0.15\% &  \\
\cmidrule{2-6}
 & \multirow{21}{*}{7B} & scala & 74,377 & 14.17\% & \multirow{21}{*}{524,708} \\
 &  & java & 68,861 & 13.12\% &  \\
 &  & c & 53,398 & 10.18\% &  \\
 &  & html & 49,576 & 9.45\% &  \\
 &  & go & 44,513 & 8.48\% &  \\
 &  & kotlin & 34,436 & 6.56\% &  \\
 &  & c\_sharp & 25,344 & 4.83\% &  \\
 &  & php & 24,442 & 4.66\% &  \\
 &  & rust & 22,102 & 4.21\% &  \\
 &  & python & 20,861 & 3.98\% &  \\
 &  & javascript & 15,167 & 2.89\% &  \\
 &  & lua & 13,309 & 2.54\% &  \\
 &  & sql & 12,755 & 2.43\% &  \\
 &  & shell & 11,726 & 2.23\% &  \\
 &  & typescript & 11,561 & 2.20\% &  \\
 &  & swift & 11,075 & 2.11\% &  \\
 &  & ruby & 10,645 & 2.03\% &  \\
 &  & css & 9,104 & 1.74\% &  \\
 &  & dockerfile & 5,755 & 1.10\% &  \\
 &  & julia & 4,849 & 0.92\% &  \\
 &  & r & 852 & 0.16\% &  \\
\bottomrule
\caption{Domain distribution of memorized texts across model families and sizes.}
\label{tab:memorized-domain-distribution}
\end{longtable}
\twocolumn
\clearpage

\end{document}